\documentclass[journal]{IEEEtran}

\usepackage{wrapfig}
\usepackage{subfigure}
\usepackage{amssymb}

 \usepackage[colorlinks=true,hyperindex=true,bookmarksnumbered,bookmarks=true]{hyperref} 
 \usepackage{multirow}
\usepackage[colorlinks=true,hyperindex=true,bookmarksnumbered,bookmarks=true]{hyperref} 
\usepackage{multirow}
\usepackage{bbold}

\usepackage[utf8]{inputenc}
\usepackage[english]{babel}

\usepackage{listings}
\usepackage{color}
\usepackage{tikz}
\usepackage{tikz-layers}
\usepackage{array}
\newcolumntype{C}[1]{>{\centering\arraybackslash}p{#1}}
\usetikzlibrary{shapes.geometric,automata,positioning,arrows,shadows,,backgrounds,calc}

\definecolor{frameColor}{RGB}{128,128,128}
\definecolor{initialStateColor}{RGB}{10,100,10}
\definecolor{impactStateColor}{RGB}{0,255,0}
\definecolor{admittanceStateColor}{RGB}{0,0,255}

\definecolor{resetStateColor}{RGB}{10,100,10}

\definecolor{stateColor}{RGB}{51,204,204}

\definecolor{controllerColor}{RGB}{10,10,255}
\definecolor{plannerColor}{RGB}{10,150,150}
\definecolor{estimatorColor}{RGB}{10,200,10}
\definecolor{robotColor}{RGB}{255,0,0}
\definecolor{modelColor}{RGB}{180,100,10}

\usepackage{kantlipsum}

\usepackage{setspace}

\usepackage{listings}
\newcommand{\quickEq}[2]{
  \begin{equation}
    \label{#1}
    {#2}
  \end{equation}
}

\usepackage{xspace}
\newcommand{\unitkHz}{~kHz\xspace}
\newcommand{\unitHz}{~Hz\xspace}
\newcommand{\unitMs}{~ms\xspace}

\newcommand{\unitVelTS}{~m/s\xspace}

\newcommand{\unitTorque}{~N$\cdot$m\xspace}

\usepackage{pifont}
\newcommand{\contactVel}{\bs{v}}
\newcommand{\nContactVel}{v_n}
\newcommand{\nContactVelR}{v_{nr}}

\newcommand{\vSpringL}{x}
\newcommand{\vSpringLd}{\dot{x}}
\newcommand{\vSpringC}{k}
\newcommand{\vDamperC}{c}

\newcommand{\pEnergy}{{{E}_p}}
\newcommand{\dEnergy}{{{E}_d}}
\newcommand{\dEnergyd}{{\dot{E}_d}}

\newcommand{\kEnergy}{{E_k}}
\newcommand{\kEnergyd}{{\dot{E}_k}}

\newcommand{\pEnergyd}{\dot{E}_p}

\newcommand{\dni}[1]{\frac{d #1}{d \nImpulse} }

\newcommand{\iim}{W}
\newcommand{\emiim}[1]{\mass_{#1}}
\newcommand{\emiimDef}[2]{\frac{1}{\transpose{#2} #1 #2}}

\newcommand{\eCoefR}{e_{\text{r}}}
\newcommand{\coefA}{e_{\text{a}}}

\newcommand{\preImpact}[1]{#1^{-}}
\newcommand{\coefR}{c_{\text{r}}}

\newcommand{\basisVec}[1]{\widehat{\bs{\bs{#1}}}}

\newcommand{\tp}[1]{{#1}_{\perp}}

\newcommand{\nVel}{\linearVelScalar_{n}}

\newcommand{\nImpulse}{{\impulseScalar_{n}}}

\newcommand{\nForce}{\forceScalar_{n}}

\newcommand{\nImpulseCE}{\impulseScalar_{nc}}

\definecolor{csrStateColor}{RGB}{255,255,153}
\definecolor{scrStateColor}{RGB}{153,204,0}
\definecolor{crStateColor}{RGB}{153,204,155}

\usepackage{stackengine}

\newcommand{\derivative}[1]{\frac{d}{dt}(#1)}

\newcommand{\inertialFrame}{O}

\newcommand{\abs}[1]{\left\vert#1\right\vert}
\newcommand{\cross}[2]{ #1 \times #2 }

\newcommand{\innerP}[2]{
  \transpose{#1}#2
}

\newcommand{\wrenchAS}[2]{S}

\usepackage{algorithmicx}
\usepackage{algorithm}
\usepackage{algpseudocode}

\newcommand{\crbGInertia}{
  {^{\text{crb}}I}
}

\newcommand{\inertiaMatrix}{I}
\newcommand{\JSIM}{M}
\newcommand{\eqInertiaMatrix}{{^{\text{eq}}I}}

\newcommand{\eInertiaTensor}{{I_{eq}}}

\newcommand{\agg}[3]{
\sum^{#2}_{#1=1}{#3}
}

\newcommand{\vc}[1]{\bs{#1}}

\newcommand{\vectorTwo}[2]{
  \begin{bmatrix}
    #1\\
    #2
\end{bmatrix}
}

\newcommand{\vectorThree}[3]{
  \begin{bmatrix}
    #1\\
    #2\\
    #3
\end{bmatrix}
}

\newcommand{\matrixTwo}[4]{
  \begin{bmatrix}
    #1 & #2\\
    #3 & #4
\end{bmatrix}
}

\newcommand{\matrixThree}[9]{
  \begin{bmatrix}
    #1 & #2 & #3\\
    #4 & #5 & #6\\
    #7 & #8 & #9
\end{bmatrix}
}

\newcommand{\linearVelScalar}{
  v
}

\newcommand{\bodyVel}[2]{
  \textcolor{bodyVelColor}{
    \vc{V}_{#1#2}
  }}

\newcommand{\bodyTV}[2]{
  \textcolor{bodyVelColor}{
    \vc{v}_{#1#2}
  }}

\newcommand{\bodyTVd}[2]{
  \textcolor{bodyVelColor}{
    \dot{\vc{v}}_{#1#2}
}}

\newcommand{\bodyRV}[2]{
  \textcolor{bodyVelColor}{
    \vc{w}_{#1#2}
}}
\newcommand{\bodyRVd}[2]{
  \textcolor{bodyVelColor}{
    \dot{\vc{w}}_{#1#2}
}}

\newcommand{\rotation}[2]{R_{#1 #2}}

\newcommand{\rotationInv}[2]{R^{\top}_{#1 #2}}

\newcommand{\translation}[2]{
  \vc{p}_{#1 #2}
}

\newcommand{\translationSkew}[2]{\skewMatrix{\vc{p}}_{#1 #2}}

\newcommand{\identityMatrix}{\mathbb{1}} 
\newcommand{\zeroMatrix}{0} 

\newcommand{\transform}[2]{
  \textcolor{geometricColorVariable}{
    g_{#1#2}
  }}

\newcommand{\twistTransformTwo}[2]{
  \textcolor{geometricColorVariable}{
    \adg{#2}{#1}
  }
}

\newcommand{\twistTransform}[2]{
  \textcolor{geometricColorVariable}{
    \adgInv{#1}{#2}
  }
}

\newcommand{\twistTransformDef}[2]{
  \adgInvDef{#1}{#2}
}

\newcommand{\bodyVelTransform}[3]{
  \textcolor{bodyVelColor}{
    \twistTransform{#1}{#2}\bodyVel{#3}{#1} + \bodyVel{#1}{#2}
  }
}

\newcommand{\adg}[2]{
  \textcolor{geometricColorVariable}{
    Ad_{g_{#1#2}}
  }
}

\newcommand{\adgInv}[2]{
  \textcolor{geometricColorVariable}{
  Ad^{-1}_{g_{#1#2}}
}}

\newcommand{\adgTrans}[2]{
  \textcolor{geometricColorVariable}{
  Ad^{\top}_{g_{#1#2}}
}}

\newcommand{\adgInvDef}[2]{
  \textcolor{geometricColorVariable}{
  \begin{bmatrix}
    \rotationInv{#1}{#2} & -\rotationInv{#1}{#2}\translationSkew{#1}{#2} \\
    \zeroMatrix & \rotationInv{#1}{#2}
\end{bmatrix}
}}

\newcommand{\geometricFTDef}[2]{
  \textcolor{geometricColorVariable}{
    \adgInvTransDef{#2}{#1}
  }
}

\newcommand{\geometricFTTwo}[2]{
  \textcolor{geometricColorVariable}{
    \adgTrans{#1}{#2}
  }
}

\newcommand{\geometricFT}[2]{
  \textcolor{geometricColorVariable}{
    \adgInvTrans{#2}{#1}
  }
}

\newcommand{\adgInvTrans}[2]{
  \textcolor{geometricColorVariable}{
  Ad^{\top}_{g^{-1}_{#1#2}}
}}

\newcommand{\adgInvTransDef}[2]{
  \textcolor{geometricColorVariable}{
  \begin{bmatrix}
    \rotation{#1}{#2} & \zeroMatrix \\
    \translationSkew{#1}{#2}\rotation{#1}{#2}  & \rotation{#1}{#2}
  \end{bmatrix}
  }
}

\newcommand{\wrench}{\bs{W}}

\newcommand{\gInertiaTransform}[3]{
  \transpose{\twistTransform{#1}{#2}}
  #3
  \twistTransform{#1}{#2}
}

\newcommand{\metaInertia}{\mathcal{I}}

\newcommand{\bodyJacobian}[2]{
  \textcolor{bodyVelColor}{
    \jacobian_{#1 #2}
  }
}

\newcommand{\desired}[1]{#1^{*}}

\newcommand{\quadratic}[2]{\frac{1}{2} #1^\top #2 #1}

\newcommand{\bs}{\boldsymbol}
\newcommand{\backfill}{\hfill\(\blacksquare\)}

\newcommand{\skewMatrix}[1]{\widehat{#1}}

\newcommand{\transpose}[1]{{#1}^\top}

\newcommand{\inverse}[1]{{#1}^{-1}}

\newcommand{\cframe}[1]{\mathcal{F}_{#1}}

\newcommand{\mass}{\text{m}}

\newcommand{\contactPoint}{\bs{p}}

\newcommand{\force}{\bs{f}}
\newcommand{\forceScalar}{f}

\newcommand{\torque}{\bs{\tau}}

\newcommand{\momentum}{\bs{h}}

\newcommand{\impactDuration}{\delta t}

\newcommand{\nAJ}{n}

\newcommand{\jump}{\Delta}

\newcommand{\com}{{\bs{c}}}

\newcommand{\impulse}{\bs{\iota}} 
\newcommand{\impulseScalar}{\iota} 

\newcommand{\jvelocities}{\mathbf{\dot{q}}}

\newcommand{\jacobian}{J}

\newtheorem{remark}{Remark}[section]

\newtheorem{Definition}{Definition}

\usepackage{mathtools}

\newcommand{\RRv}[1]{\mathbb{R}^{#1}}
\newcommand{\RRm}[2]{\mathbb{R}^{#1 \times #2}}

\definecolor{dkgreen}{rgb}{0,0.6,0}
\definecolor{gray}{rgb}{0.5,0.5,0.5}
\definecolor{mauve}{rgb}{0.58,0,0.82}

\NewDocumentCommand{\cpp}{v}{%
\texttt{\textcolor{blue}{#1}}%
}

\NewDocumentCommand{\secRef}{v}{%
Sec.~\ref{#1}}

\NewDocumentCommand{\remarkRef}{v}{%
Remark.~\ref{#1}}

\NewDocumentCommand{\tableRef}{v}{%
Table.~\ref{#1}}

\NewDocumentCommand{\algRef}{v}{%
Algorithm.~\ref{#1}}

\NewDocumentCommand{\lemmaRef}{v}{%
Lemma.~\ref{#1}}

\NewDocumentCommand{\theoremRef}{v}{%
Theorem.~\ref{#1}}

\NewDocumentCommand{\figRef}{v}{%
  Fig.~\ref{#1}}

\NewDocumentCommand{\appRef}{v}{%
  Appendix~\ref{#1}}

\NewDocumentCommand{\lineRef}{v}{%
line.~\ref{#1}}

\NewDocumentCommand\orderedTwoS{mm}%
  {$<$#1,#2$>$}
\NewDocumentCommand\orderedThreeS{mmm}%
  {$<$#1,#2,#3$>$}

\definecolor{svaColor}{RGB}{7, 160, 2}
\definecolor{uncertainColor}{RGB}{255, 0, 0}

\definecolor{spatialVelColor}{RGB}{189, 38, 96}
\definecolor{geometricColor}{RGB}{66, 126, 245}
\definecolor{bodyVelColor}{RGB}{13, 191, 191}

\colorlet{svaColorVariable}{svaColor}
\colorlet{geometricColorVariable}{geometricColor}

\newif\ifBlockComment
\BlockCommentfalse 

 \usepackage{caption}
\begin{document}
%
\title{
  On Inverse Inertia Matrix and Contact-Force Model for Robotic Manipulators at Normal Impacts
}
%
%
%
\author{
Yuquan Wang, 
Niels Dehio, and 
Abderrahmane Kheddar,~\IEEEmembership{Fellow,~IEEE}
 \thanks{This work is in part supported by the Research Project I.AM. through the European Union H2020 program (GA 871899).}
\thanks{Y. Wang, N. Dehio and A. Kheddar are with the CNRS-University of Montpellier LIRMM Interactive Digital Humans group, Montpellier, France. {\tt\small yuquan.wang,niels.dehio,kheddar@lirmm.fr}}
\thanks{A. Kheddar is also with the CNRS-AIST Joint Robotics Laboratory, IRL, Tsukuba, Japan.} 
}

\maketitle

\begin{abstract}
    State-of-the-art impact dynamics models either apply for free-flying objects or do not account that a robotic manipulator is commonly high-stiffness controlled.
  Thus, we lack tailor-made models for manipulators mounted on a fixed base. Focusing on orthogonal point-to-surface impacts (no tangential velocities), we revisit two main elements of an impact dynamics model: the contact-force model and the inverse inertia matrix. We collect contact-force measurements by impacting a 7~DOF Panda robot against a sensorized rigid environment with various joint configurations and velocities. Evaluating the measurements from 150 trials, the best model-to-data matching suggests a viscoelastic contact-force model and computing the inverse inertia matrix assuming the robot is a composite-rigid body.
\end{abstract}

\begin{IEEEkeywords}
Contact modeling, impact-awareness, dynamics.
\end{IEEEkeywords}

\IEEEpeerreviewmaketitle

\section{Introduction}

\IEEEPARstart{W}{hen}
a high-stiffness controlled robot impacts rigid surfaces, the robot's joint velocities and torques values will change instantly, within a fraction to dozens of milliseconds. If not restricted to their tolerable range, such state jumps may have severe consequences, up to damaging the robot's hardware and/or its surroundings. Therefore, close-to-zero contact velocity is generally planned to avoid impacts, e.g.,~\cite{Winkler2018RAL}. This workaround, however, prevents implementing impact-based tasks like hammering, dynamic loco-manipulations, or heavy box swift grabbing, to name just a few. A reliable impact dynamics model would allow the robot controller to regulate the contact velocities according to predicted post-impact states. 

State-of-the-art robot controllers, e.g.,~\cite{grizzle2014automatica,siciliano2016springer,wang2020ijrr}
predicts impact subsequent impulses using the algebraic equations developed in the late 1980s~\cite{zheng1985mathematical}.
Impact studies such in~\cite{stronge2000book,jia2017ijrr,halm2019rss} proposed more refined models, yet most of them assume impacts between two free-flying bodies~\cite{jia2019ijrr}.

\begin{figure}[tbp!]
  \centering
  \includegraphics[width=\columnwidth]{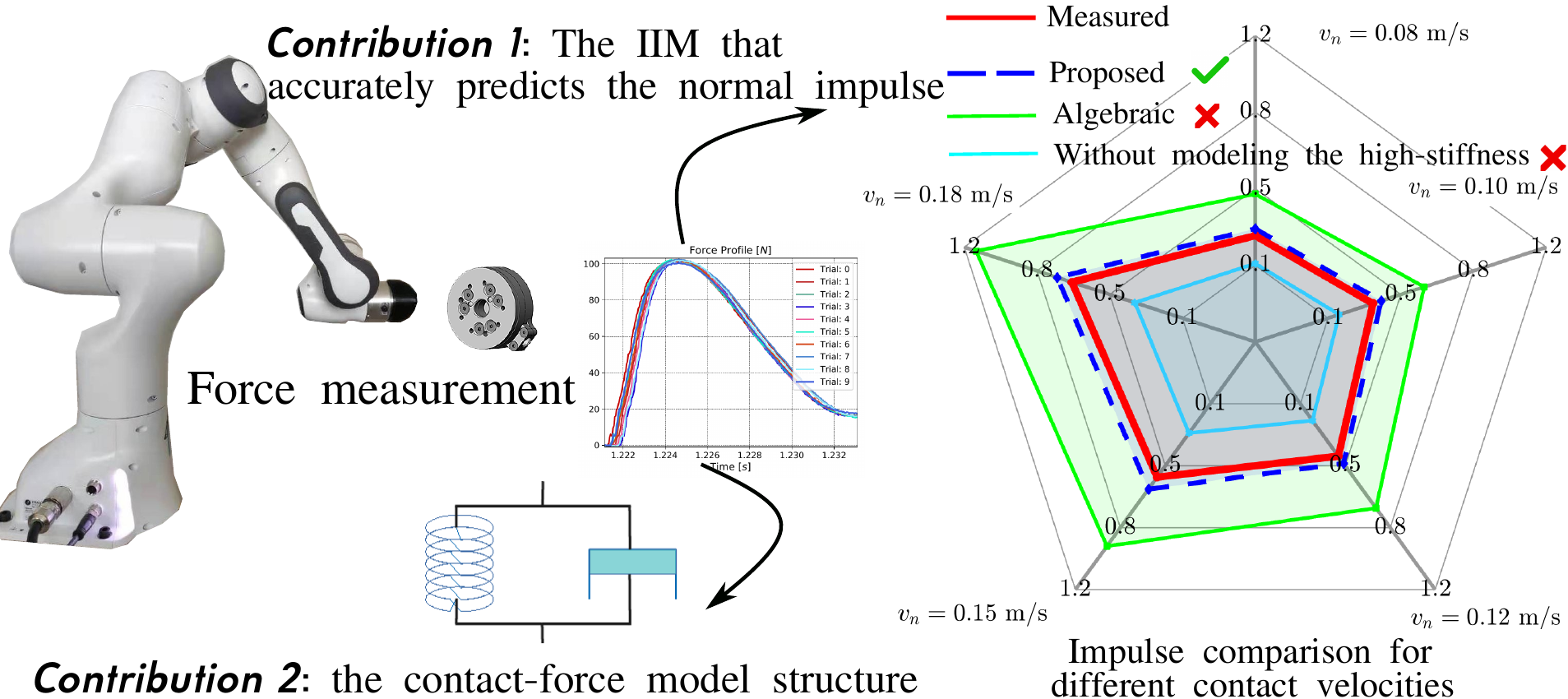}
  \caption{
    According to measured contact forces, we conclude for high-stiffness controlled manipulators: (1) computing IIM assuming the robot is a composite-rigid body;  (2) the structure of the contact-force model is a parallel connection of a virtual spring and a dashpot.
}
\label{fig:comparison}
\end{figure}

Under active-continuous joint control, fixed-base manipulators will not bounce as a free-floating mass would have done. In front of the difficulty of having reliable and sound impact predictions using textbook models, we investigate the reasons for this shortcoming.
Therefore, we devised a benchmark study using the Panda robot in a well-calibrated and instrumented environment, see~\figRef{fig:comparison}.

Predicting the post-impact states relies on (at least) two essential ingredients:
(i) a good estimate or the computation of the
task-space velocity-to-impulse mapping, i.e., the \emph{inverse inertia matrix} (IIM), and
(ii) a  \emph{contact-force model}.

A well-defined IIM~\cite{stronge2000book} allows predicting
(a) the post-impact contact mode (i.e., sliding or sticking), 
(b) the stable slip direction (i.e., if the tangential contact velocity converged to an invariant direction), and 
(c) the contact velocity given an impulse (during an impact event).  
We theoretically derive the IIM based on different assumptions:
\begin{enumerate}
\item inverse of the joint-space inertia matrix~\cite{grizzle2014automatica,siciliano2016springer,zheng1985mathematical,aouaj2021icra};
\item \label{item:crb_option} considering the joints with high-stiffness, hence treating the robot as a composite-rigid body (CRB); 
\item applying the joint motion constraint (without considering high-stiffness joints)~\cite{lankarani2000poisson,khulief2013modeling,aghili2017impact}.
\end{enumerate}
According to the data collected from 150 impact experiments,
our findings suggest that we can predict the normal impulse most accurately with 
option~(2).

Different contact-force models lead to drastically different timing of the events, e.g., \emph{the end of compression} or, more importantly, \emph{the end of restitution}, which 
determines the impulse and post-impact velocities~\cite{jia2017ijrr,halm2019rss}. 
Based on the measured contact forces, we found the deformation-rate-dependent (viscoelastic) compliance is not negligible, i.e., the
pure elastic contact-force model 
for two free-flying bodies~\cite{jia2017ijrr,pashah2008prediction}~\cite[Chapter~2]{stronge2000book} is not applicable for high-stiffness controlled manipulators. 
The viscoelasticity enables representing
energy-dissipation by damping and indicates
a decreasing estimated coefficient of restitution (COR) when the contact velocity increases~\cite{stronge2000book}. 
In all the experiments, the estimated COR is smaller than the material-dependent COR.

To summarize, for the impacts conducted by a high-stiffness controlled manipulator, our findings are:
\begin{enumerate}
\item option (2) is the measurement-consistent way to compute the IIM;
  see the derivation in~\secRef{sec:crb} and the validation in~\secRef{sec:validate_iim}. 
\item the contact-force model is viscoelastic, see~\secRef{sec:two_phases}.\end{enumerate}
Our analysis is based on the following assumptions:
\begin{itemize}
\item The impact force is large compared to body forces and centripetal inertial terms.
Other forces remain constant during impact~\cite[Chapter~8.1.1]{stronge2000book}.
\item Point contact. The  contact area is negligibly small compared to the robot dimensions \cite{chatterjee1998new}. 
\item Negligible impact-induced  contact moments~\cite{stronge2000book,chatterjee1998new}.
\item Normal-to-surface impacts (tangential impulse is negligible)~\cite{stronge2000book,jia2017ijrr}.
\item The fixed-base fully-actuated robot under high sampling rate, e.g., 1000\unitHz position/velocity control. 
\item The impacting bodies in our experiments are locally deformable for the chosen range of contact velocities between $0.08-0.18$\unitVelTS.
\end{itemize}
\section{Related work}

The IIM is essential to predict post-impact states in rigid-body dynamics, e.g.,~\cite{zheng1985mathematical,wang1992two,hurmuzlu1994ijrr,lankarani2000poisson,stronge2000book,khulief2013modeling,aghili2017impact,jia2017ijrr,jia2019ijrr,halm2019rss}. There are impact computations dealing specifically with non-articulated objects, e.g., \cite{wang1992two,stronge2000book,jia2017ijrr,jia2019ijrr,halm2019rss},
or articulated linkages yet without accounting for the controller behavior, e.g., the under-actuated pendulums studied by~\cite{lankarani2000poisson,khulief2013modeling,ganguly2020ijrr} with experiments and by \cite{hurmuzlu1994ijrr,aghili2017impact} in simulation. According to our benchmark experiments, both situations
do not accurately predict the impulse (between a high-stiffness controlled manipulator and the rigid environment). Our IIM computation accounts for both the robot joint motion constraints~\cite{hurmuzlu1994ijrr,lankarani2000poisson,khulief2013modeling,aghili2017impact,ganguly2020ijrr}; and the high-stiffness aspect by assuming the robot is a composite-rigid body during the short time span of an impact.

The mass-spring-damper model is widely adopted to describe the normal contact force~\cite{pashah2008prediction}.  
Elasticity is commonly
used for impacts between free-flying bodies~\cite{stronge2000book,jia2017ijrr,dehio2021icra}. 
However, pure elasticity contradicts the fact that our measured peak force is not in phase with the compression. According to our benchmark experiments, the viscoelastic models reported by Stronge in~\cite{stronge2000book} determine the impact events more accurately for robot under high-stiffness joint control. 
There are many continuous point-contact force models~\cite{skrinjar2018review}. However, the model parameters might change regarding different control modes, and
the inclusion of the contact-force model in the equations of motion may result in computational inefficiency or failure of numerical integration routines~\cite{lankarani2000poisson}. Thus, we focus on identifying the model structure, i.e., linear spring and nonlinear dashpot,
without explicitly comparing the best-fit model from all the candidates.

\section{Notations}
We define the IIM and introduce commonly-applied computations in the rest of the paper.
\begin{Definition}
Assuming  impact does not generate impulsive moment, we evaluate the contact velocity jump  $\jump \contactVel \in \RRv{3}$ during the impact via the \emph{inverse inertia matrix} $\iim \in \RRm{3}{3}$  and the impulse $\impulse\in\RRv{3}$ \cite{stronge2000book}:
\quickEq{eq:iim_def}{
\jump \contactVel= \iim \impulse.
}
\end{Definition}
We borrow the notations from the book by Murray~\emph{et al}.~\cite{murray1994book}. To ease the reading, we mark the body velocities and associated Jacobians in cyan, e.g., the body velocity of link $i$ (with respect to the inertial frame $\cframe{\inertialFrame}$ and represented in the link frame $\cframe{i}$)
is\footnote{Since we use only body velocities, we omit the subscript $^b$; $\bodyVel{\inertialFrame}{i}$ is noted by $\vc{V}^b_{{\inertialFrame}{i}}$ in \cite{murray1994book}. } $\bodyVel{\inertialFrame}{i} \in \RRv{6}$, which concatenates the linear velocity  $\bodyTV{\inertialFrame}{i} \in \RRv{3}$ and the rotational velocity $\bodyRV{\inertialFrame}{i} \in \RRv{3}$:
$$
\bodyVel{\inertialFrame}{i} =
\vectorTwo{
  \bodyTV{\inertialFrame}{i}
}{
  \bodyRV{\inertialFrame}{i}
}.
$$
We mark the adjoint transform and its expansions by blue color.
In the rest of the paper, we apply the following velocity, wrench and inertia transforms:  \\
(1): Transform the \emph{body velocity} $\bodyVel{\inertialFrame}{\contactPoint}$ to  frame $\cframe{\com}$:
\quickEq{eq:vel_transform}{
  \twistTransform{\contactPoint}{\com}\bodyVel{\inertialFrame}{\contactPoint} = \twistTransformDef{\contactPoint}{\com}\bodyVel{\inertialFrame}{\contactPoint},
}
where $\rotation{\contactPoint}{\com} \in \RRm{3}{3}, \translation{\contactPoint}{\com} \in \RRv{3}$ denote the relative rotation and the relative translation, respectively. The skew-symmetric matrix  $\translationSkew{\contactPoint}{\com}$ converts the cross product by matrix multiplication. \\
(2): Transform the wrench $\wrench_e \in \RRv{6}$ represented in frame $\cframe{e}$ to frame $\cframe{\com}$:
\quickEq{eq:w_transform}{
  \geometricFT{e}{\com}\wrench_e = \geometricFTDef{e}{\com}\wrench_e.
}
Note that momentum  transform is the same as \eqref{eq:w_transform}  \cite{featherstone2014book}. \\
(3): Transform the inertia matrix $\inertiaMatrix_i \in \RRm{6}{6}$ represented in frame $\cframe{i}$ to frame $\cframe{\com}$:
\quickEq{eq:m_transform}{
  \gInertiaTransform{\com}{i}{\inertiaMatrix_i}.
}
\section{The Inverse Inertia Matrix}
\label{sec:iim}
We studied three ways to compute $\iim$:
(1) inverse of the generalized (joint-space) momentum in~\secRef{sec:projection_iim},
(2) assuming the robot is a composite-rigid body (CRB) in~\secRef{sec:crb},
(3) without considering joints' high-stiffness in~\secRef{sec:crb_flexibility}.
We leave the details of computing the normal impulse with a particular IIM in~\appRef{sec:nImpulse_calc}.

\subsection{Projection Approach}
\label{sec:projection_iim}
From the principle of kinetic energy conservation, the body velocity $\bodyVel{\inertialFrame}{\contactPoint} \in \RRv{6}$ and the equivalent inertia matrix $\eInertiaTensor \in \RRm{6}{6}$ produce the same amount of kinetic energy as the joint space inertia matrix $\JSIM \in \RRm{\nAJ}{\nAJ}$ and velocities $\jvelocities \in \RRv{\nAJ}$
\quickEq{eq:kinetic_energy_conservation}{
  \quadratic{\jvelocities}{\JSIM} 
    =  \quadratic{\bodyVel{\inertialFrame}{\contactPoint}}{
    \underbrace{\inverse{(\jacobian\inverse{\JSIM} \transpose{\jacobian})}}_{\eInertiaTensor}
  }.
}
We denote $\bodyJacobian{\inertialFrame}{\contactPoint}$ by $\jacobian$ in (\ref{eq:kinetic_energy_conservation}-\ref{eq:option_em}) to avoid lengthy notations. The equality~\eqref{eq:kinetic_energy_conservation} leads to two options\footnote{We reserve the first 3 rows for translational velocity of the Jacobian $\jacobian \in \RRm{6}{\nAJ}$. If the notations are in line with the book by Featherstone \cite{featherstone2014book}, we need to take the lower-right corner.} to compute the impulse:
\begin{align}
  \label{eq:option_gm_iim}
  \iim_{gm}:~3\times3\text{ upper-left corner of}&~\jacobian\inverse{\JSIM} \transpose{\jacobian}  \cite{lankarani2000poisson,khulief2013modeling};\\
  \label{eq:option_em}
  \mass_{em}:~3\times3\text{ upper-left corner of}&~\inverse{(\jacobian\inverse{\JSIM} \transpose{\jacobian})}.
\end{align}

The first option is substituting $\iim_{gm}$ into the well-known procedure \eqref{eq:zImpulseCE} in \appRef{sec:nImpulse_calc}. The other option is
the  algebriac equation  \cite{grizzle2014automatica,siciliano2016springer,zheng1985mathematical} that computes  as:
\quickEq{eq:sota_gm_iim}{
  \impulse =  (1 + \eCoefR)\mass_{em} \preImpact{\contactVel} = \mass_{em} \jump \bodyTV{\inertialFrame}{\contactPoint},
}
where   $\preImpact{\contactVel} \in \RRv{3}$ denotes the (pre-impact) contact velocity and the coefficient of restitution belongs to $ \eCoefR \in [0, 1]$.
\subsection{Composite-rigid-body approach}
\label{sec:crb}
Let $\force_{\contactPoint} \in \RRv{3}$ be the external force applied at contact point~$\contactPoint$. The wrench $\wrench_{i} \in \RRv{6}$ at the $i$th link writes:
\quickEq{eq:induced_wrench}{
  \wrench_{i} = \geometricFT{\contactPoint}{i}\vectorTwo{\force_{\contactPoint}}{0}.
}
Given the mass $\mass_i \in \RRv{}$ and the moment of inertia $\metaInertia_i \in \RRm{3}{3}$, Newton-Euler's equation in the body coordinates writes:
\quickEq{eq:newton_euler_body}{
  \matrixTwo{\mass_i \identityMatrix }{\zeroMatrix}{\zeroMatrix}{\metaInertia_i}
  \vectorTwo{\bodyTVd{\inertialFrame}{i}}{\bodyRVd{\inertialFrame}{i}}
  +
  \vectorTwo{\cross{\bodyRV{\inertialFrame}{i}}{\mass_i\bodyTV{\inertialFrame}{i}}}{\cross{\bodyRV{\inertialFrame}{i}}{\metaInertia_i\bodyRV{\inertialFrame}{i}}}
  = \wrench_{i},
}
where $ \identityMatrix \in \RRm{3}{3}$ is an identity matrix.
Substituting~\eqref{eq:induced_wrench} into~\eqref{eq:newton_euler_body}, we compute the momentum jump $\jump \momentum_i$ by integrating~\eqref{eq:newton_euler_body} over the impact duration~$\impactDuration$
\quickEq{eq:link_momentum_jump}{
\jump \momentum_i = \matrixTwo{\mass_i  \identityMatrix }{\zeroMatrix}{\zeroMatrix}{\metaInertia_i}
\vectorTwo{\jump \bodyTV{\inertialFrame}{i}}{\jump \bodyRV{\inertialFrame}{i}}
= \geometricFT{\contactPoint}{i}\vectorTwo{\impulse}{0},
}
where the cross product in \eqref{eq:newton_euler_body} vanishes as the impact force is large w.r.t the centripetal inertial terms~\cite[Chapter~8.1.1]{stronge2000book}.

In order to compute the impact-induced whole-body momentum jump, 
we transform each $\jump \momentum_i$ to the centroidal frame $\cframe{\com}$ according to \eqref{eq:w_transform} and  aggregate the transformed momentum jump of  all the links:
\quickEq{eq:cm_jump}{
\jump \momentum  = \agg{i}{\nAJ}{
  \geometricFT{i}{\com}
  \jump \momentum_i}.  
}
The jump of the average velocity \cite[Eq.~24]{orin2013auro} defined in the centroidal frame writes: 
\quickEq{eq:comd_jump}{
  \jump \bodyVel{\inertialFrame}{\com} = \inverse{\crbGInertia} \jump \momentum,
}
where the centroidal inertia $\crbGInertia \in \RRm{6}{6}$ is similar to~\cite[Eq.~22]{orin2013auro}. The relation between $\crbGInertia$ and the inertia matrix $\inertiaMatrix_i$ of a specific link $i$ is explained in~\appRef{app:ci}.

We re-write the contact point body velocity $\bodyVel{\inertialFrame}{\contactPoint} \in \RRv{6}$ relative to the centroidal frame $\cframe{\com}$ (which is between the inertial frame $\cframe{\inertialFrame}$ and the contact point frame $\cframe{\contactPoint}$) 
according to  \cite[Proposition~2.15]{murray1994book}:
\quickEq{eq:contactVel_exact}{
  \bodyVel{\inertialFrame}{\contactPoint} = \bodyVelTransform{\com}{\contactPoint}{\inertialFrame}.
}
The CRB assumption leads to the relative velocity between the centroidal frame and the contact point is zero ($\bodyVel{\com}{\contactPoint} = 0$) such that we can approximate:
\quickEq{eq:contactVel_transform}{
\bodyVel{\inertialFrame}{\contactPoint} \approx \twistTransform{\com}{\contactPoint} \bodyVel{\inertialFrame}{\com}, 
}
which amounts to transforming the average velocity $\bodyVel{\inertialFrame}{\com}$ to the contact point according to the velocity transform \eqref{eq:vel_transform}. 
Substituting $\jump \bodyVel{\inertialFrame}{\com}$ from \eqref{eq:comd_jump} into \eqref{eq:contactVel_transform}, the contact point velocity jump induced by the external impulse is:
\quickEq{eq:iim_full}{
\begin{aligned}
  &\jump \bodyVel{\inertialFrame}{\contactPoint} \approx \twistTransform{\com}{\contactPoint} \inverse{\crbGInertia} \jump \momentum \\ 
  & = \twistTransform{\com}{\contactPoint} \inverse{\crbGInertia}
  \agg{i}{\nAJ}{
    \underbrace{\geometricFT{i}{\com}
   \jump \momentum_i}}_{\text{Momentum transform \eqref{eq:w_transform} from $\cframe{i}$ to $\cframe{\com}$}}  \\
 & = \twistTransform{\com}{\contactPoint} \inverse{\crbGInertia}
  \agg{i}{\nAJ}{
   \geometricFT{i}{\com}
   \underbrace{\geometricFT{\contactPoint}{i}\vectorTwo{\impulse}{0}}_{\text{According to \eqref{eq:link_momentum_jump}}}
 } \\
 & = \twistTransform{\com}{\contactPoint} \inverse{\crbGInertia}
  \agg{i}{\nAJ}{
    \geometricFT{\contactPoint}{\com}\vectorTwo{\impulse}{0}
  }\\
  & =
  \underbrace{\twistTransform{\com}{\contactPoint} \inverse{\crbGInertia}\geometricFT{\contactPoint}{\com}}_{\text{See Remark~\ref{remark:crb}}}
  \vectorTwo{\impulse}{0} \\
  & =
  \twistTransformDef{\com}{\contactPoint}
  \matrixTwo{\frac{1}{\mass}{\identityMatrix}}{0}{0}{\inverse{\metaInertia}}
  \geometricFTDef{\contactPoint}{\com}
  \vectorTwo{\impulse}{0}.
\end{aligned}
}
Given $\bodyVel{\inertialFrame}{\contactPoint} = \vectorTwo{\bodyTV{\inertialFrame}{\contactPoint}}{\bodyRV{\inertialFrame}{\contactPoint}}$, we extract the translation from~\eqref{eq:iim_full} and obtain the inverse inertia matrix: 
\quickEq{eq:iim_inertia_def}{
  \jump\bodyTV{\inertialFrame}{\contactPoint}  =
  \underbrace{(\frac{\identityMatrix_{3 \times 3}}{\mass} - \rotationInv{\com}{\contactPoint}\translationSkew{\com}{\contactPoint} \inverse{\metaInertia} \translationSkew{\com}{\contactPoint}\rotation{\com}{\contactPoint} )}_{\iim}
  \impulse,
}
where $\mass \in \RRv{}$ is the robot's total mass, $\metaInertia \in \RRm{3}{3}$ is the  moment of inertia of $\crbGInertia$.

\begin{remark}
  \label{remark:crb}
  We left multiply $\gInertiaTransform{\contactPoint}{\com}{\crbGInertia}$ to the following intermediate step from  \eqref{eq:iim_full}: 
  \quickEq{eq:def_I_eq}{
  \begin{aligned}
  \jump \bodyVel{\inertialFrame}{\contactPoint} & \approx \twistTransform{\com}{\contactPoint} \inverse{\crbGInertia}\geometricFT{\contactPoint}{\com} 
  \vectorTwo{\impulse}{0}, \\
  \underbrace{\gInertiaTransform{\contactPoint}{\com}{\crbGInertia}}_{\text{Equivalent inertia matrix: $\eqInertiaMatrix_{\contactPoint}$}}
  \jump\bodyVel{\inertialFrame}{\contactPoint} & \approx \vectorTwo{\impulse}{0}.
  \end{aligned}
  }
  According to the inertia transform \eqref{eq:m_transform}, the equivalent inertia matrix at the contact point $\eqInertiaMatrix_{\contactPoint}$ amounts to transforming the centroidal inertia matrix $\crbGInertia$ to the contact point $\contactPoint$.
  \backfill
\end{remark}

\begin{remark}
\label{remark:inertia}
  We denote the inertia matrix of the link $i$ as $\inertiaMatrix_i \in \RRm{6}{6}$. 
  The \emph{contribution} of $\inertiaMatrix_{i}$  to  $\inverse{\eqInertiaMatrix}_{\contactPoint}$ is: 
  \quickEq{eq:contribution_im}{
    \twistTransformTwo{i}{\contactPoint} 
    \inverse{\inertiaMatrix}_i
    \transpose{\twistTransformTwo{i}{\contactPoint}},
  }
  where $\transform{\contactPoint}{i} \in SE(3)$ is the transform from the contact point frame $\cframe{\contactPoint}$ to 
  the end-effector frame $\cframe{i}$; see~\appRef{sec:inertia_proof}.
  \backfill
\end{remark}

\subsection{Without considering high-stiffness}
\label{sec:crb_flexibility}
Now suppose the relative velocity $\bodyVel{\com}{\contactPoint} \neq 0$, we denote the incremental change of $\iim$ compared to \eqref{eq:iim_inertia_def} by
\quickEq{eq:crb_fb_iim}{
  \tilde{\iim} = \iim + \iim_{\text{flexibility}}.
}
According to \eqref{eq:contactVel_exact}, we compute $\bodyVel{\com}{\contactPoint}$ as:
\quickEq{eq:internal_vel}{
  \begin{aligned}
    \bodyVel{\com}{\contactPoint} & = \bodyVel{\inertialFrame}{\contactPoint} - \twistTransform{\com}{\contactPoint}\bodyVel{\inertialFrame}{\com} 
    = \underbrace{(\bodyJacobian{\inertialFrame}{\contactPoint} - \twistTransform{\com}{\contactPoint}\bodyJacobian{\inertialFrame}{\com})}
    _{ \bodyJacobian{\com}{\contactPoint} \in \RRm{6}{\nAJ}}
      \jvelocities.  
  \end{aligned}
}

To compute the  impulse that induces  $\jump \bodyVel{\com}{\contactPoint}$, we need the joint velocity jump as the intermediate variable. Integrating the equations of motion at the moment of the impact \cite{zheng1985mathematical,wang2019rss}, we obtain $\JSIM \jump \jvelocities = \transpose{\bodyJacobian{\inertialFrame}{\contactPoint}} \vectorTwo{\impulse}{0}$. Hence, $\jump \jvelocities$ writes
\quickEq{eq:jv_jump_def}{
  \jump \jvelocities = \inverse{\JSIM} \transpose{\bodyJacobian{\inertialFrame}{\contactPoint}}
\vectorTwo{\impulse}{0}.
}
Substituting \eqref{eq:jv_jump_def} into \eqref{eq:internal_vel}:
$$
\jump \bodyVel{\com}{\contactPoint} = \bodyJacobian{\com}{\contactPoint} \jump \jvelocities =
\bodyJacobian{\com}{\contactPoint}
\inverse{\JSIM} \transpose{\bodyJacobian{\inertialFrame}{\contactPoint}}\vectorTwo{\impulse}{0}.
$$
Similarly to the derivation of \eqref{eq:iim_inertia_def},  $\iim_{\text{flexibility}}$ is the $3\times3$ upper-left corner of $\bodyJacobian{\com}{\contactPoint} \inverse{\JSIM} \transpose{\bodyJacobian{\inertialFrame}{\contactPoint}}$.
\section{The contact force model}
\label{sec:two_phases}

The nonlinear viscoelastic model in~\secRef{sec:contact_force_model}, can generate measurement-consistent contact forces. It predicts a decreasing COR if the contact velocity increases, see~\secRef{sec:cor_analysis}.
It also models the energy loss via the dissipated energy and the non-zero potential energy at the end of restitution, see~\secRef{sec:energy}.

\subsection{Contact-force model}
\label{sec:contact_force_model}
We use the viscoelastic model in~\cite[Sec.~5.1.2]{stronge2000book} at contact point $\contactPoint$. Let $\vSpringL$ be the normal relative deformation, see~\figRef{fig:model_illustration}, we choose the local coordinate frame such that the initial normal contact velocity $\preImpact{\nVel} \in \RRv{}$ is negative: $\dot{x}_0 =  \preImpact{\nVel} < 0$. The normal contact-force $\nForce\in\RRv{}$ is the derivative of the normal impulse~$\nImpulse\in\RRv{}$:
\quickEq{eq:normal_impact_force}{
  \derivative{\nImpulse} = \nForce = - \vSpringC\vSpringL - \vDamperC \abs{\vSpringL}\vSpringLd =
  \vDamperC \vSpringL\vSpringLd - \vSpringC\vSpringL,
}
where the positive scalars~$\vDamperC$ and~$\vSpringC$ denote the dashpot coefficient and the spring constant.

\subsection{Coefficient of restitution}
\label{sec:cor_analysis}
Using an impact model with the COR, the velocity when the restitution ends is
\quickEq{eq:def_contactVel_RE}{
  \nContactVelR = - \eCoefR \preImpact{\nContactVel}.
}
The deformation $\vSpringL$ keeps negative $\vSpringL < 0$ during the impact but does not restore to (initial) zero by the end of the restitution phase, see \figRef{fig:estimation} and \secRef{sec:validation}. Yet, at the end of the restitution, the contact force is almost nil. Equating \eqref{eq:normal_impact_force} to zero leads
$$
\vDamperC \vSpringLd = \vSpringC \Rightarrow  \nContactVelR  = \vSpringLd = \frac{\vSpringC}{\vDamperC}, 
$$
and substituting $\nContactVelR$ from \eqref{eq:def_contactVel_RE}, we conclude another expression of COR:
\quickEq{eq:coe_property}{
  \eCoefR  = - \frac{\vSpringC}{\vDamperC }\frac{1}{\preImpact{\nContactVel}}.
}
Therefore, if $\frac{\vSpringC}{\vDamperC }$ is constant, $\eCoefR$ decreases if the pre-impact contact velocity  $\preImpact{\nContactVel}$ increases. 

\subsection{Energy consistency}
\label{sec:energy}
At any instant of the impact process, the sum of the kinetic energy $\kEnergy$, the spring-stored potential energy
\quickEq{eq:potential_energy}{
  \pEnergy = \frac{1}{2}\vSpringC \vSpringL^2,
}
and the dashpot dissipated energy
\quickEq{eq:dissipated_energy}{
\dEnergy = \int  - \vDamperC \vSpringL\vSpringLd dx =
\int -\vDamperC \vSpringL\vSpringLd \frac{dx}{dt}dt = \int -\vDamperC \vSpringL\vSpringLd^2 dt
}
is always equal to the initial kinetic energy:
\quickEq{eq:energy_balance}{
\kEnergy(t_0) =\frac{1}{2}\mass {\preImpact{\nVel}}^2 = \kEnergy + \pEnergy + \dEnergy.
}
We can assess \eqref{eq:energy_balance} by checking its derivative: $\pEnergyd = \vSpringC \vSpringL\vSpringLd$, $\dEnergyd =  -\vDamperC \vSpringL\vSpringLd^2$ and $\kEnergyd = \mass \vSpringLd \ddot{\vSpringL} = \vSpringLd \nForce = \vDamperC \vSpringL\vSpringLd^2 -  \vSpringC \vSpringL\vSpringLd$, we find:
$$
\kEnergyd + \pEnergyd + \dEnergyd = 0.
$$
At the end of the impact process, the energy loss includes the remaining potential energy $\pEnergy$ and the dissipated energy $\dEnergy$.

\section{Data acquisition}
\label{sec:robot_configuration}
\begin{figure}[htp!]
  \begin{tabular}{C{0.5\textwidth}}
    \subfigure[b][Configuration one] {
    \resizebox{0.15\textwidth}{!}{%
    \begin{tikzpicture}
      \label{fig:impact_one}
      \begin{scope}
        \node[anchor=south west,inner sep=0] (image) at (0,0) {
          \includegraphics[width=3cm]{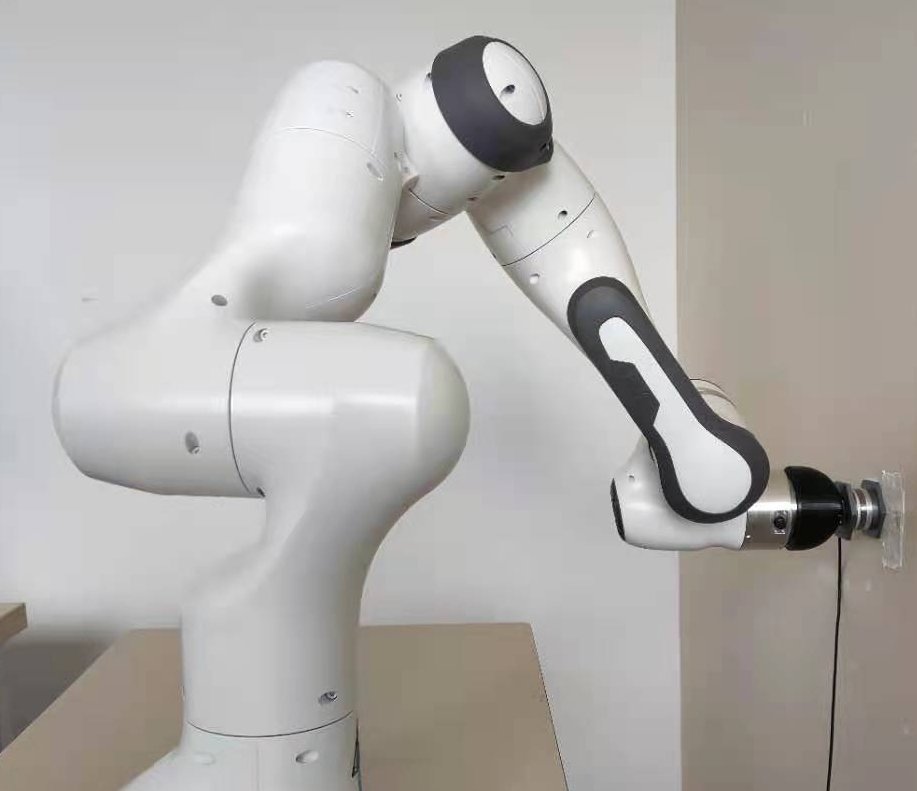}};
      \end{scope}
    \end{tikzpicture} 
    }
    }
    \subfigure[b][Configuration two] {
    \resizebox{0.135\textwidth}{!}{%
    \begin{tikzpicture}
      \label{fig:impact_two}
      \begin{scope}
        \node[anchor=south west,inner sep=0] (image) at (0,0) {
          \includegraphics[width=3cm]{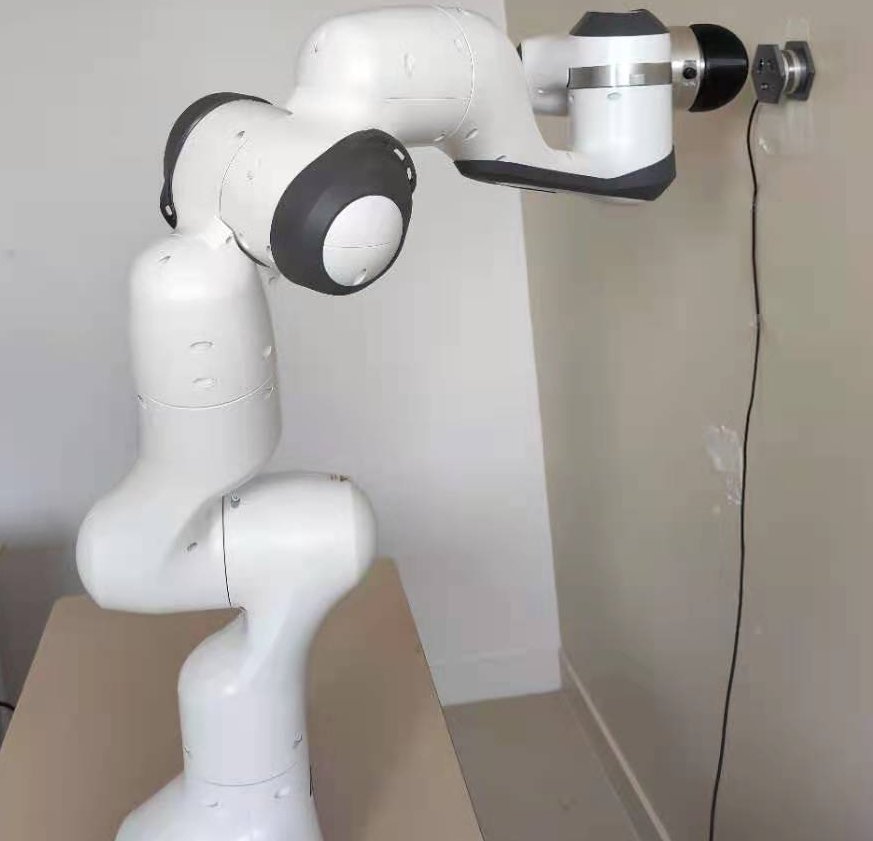}};
      \end{scope}
    \end{tikzpicture} 
    }
    }    
    \subfigure[b][Configuration three] {
    \resizebox{0.15\textwidth}{!}{%
    \begin{tikzpicture}
      \label{fig:impact_three}
      \begin{scope}
        \node[anchor=south west,inner sep=0] (image) at (0,0) {
          \includegraphics[width=3cm]{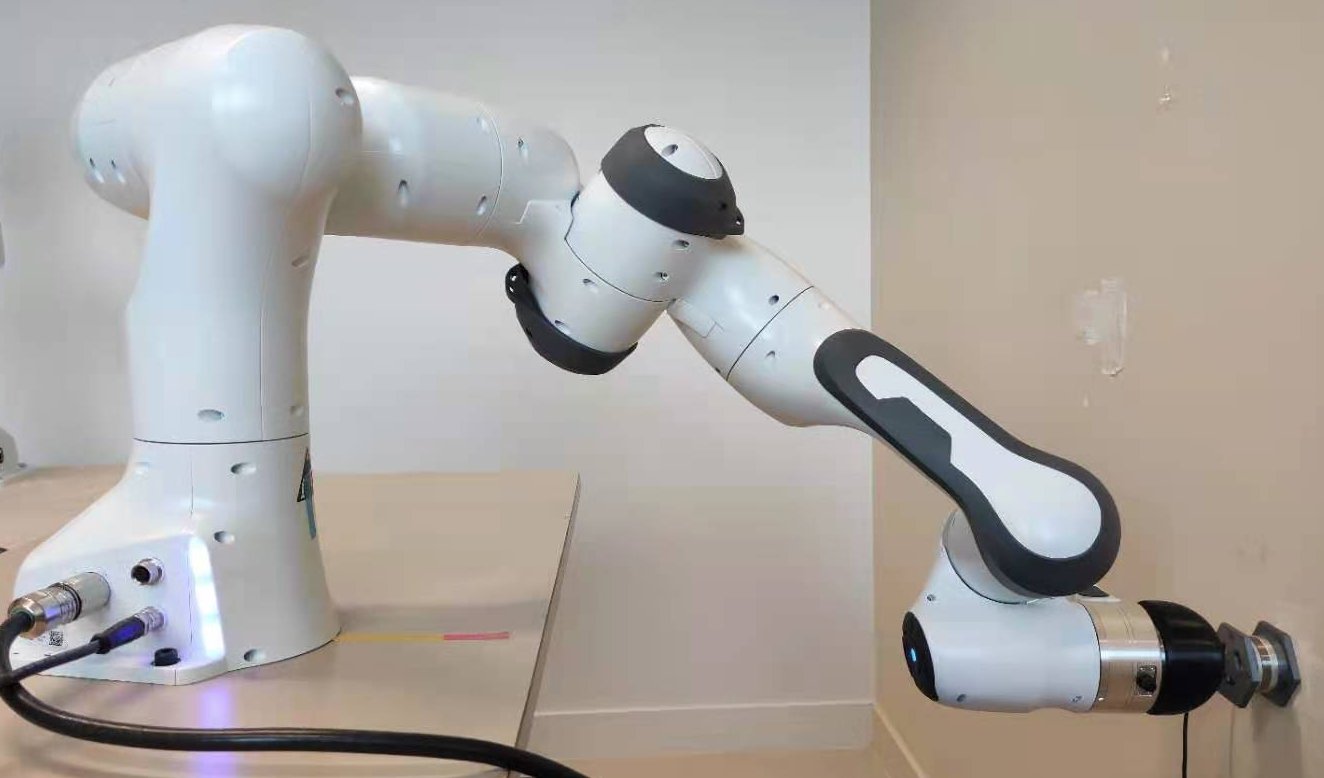}};
      \end{scope}
    \end{tikzpicture} 
    }
    }
  \end{tabular}
  \caption{
    The Panda robot impacted with three unique configurations with various reference contact velocities. 
  }
  \label{fig:impact_configurations}
\end{figure}

To keep the point contact assumption, a custom-made 3D-printed semi-spherical rigid piece is mounted on the end-effector of the 7 DOF panda robot. The robot is controlled in velocity command to impact an ATI-mini45 force-torque sensor fixed at different spots of a rigid wall. The controller loop runs at $1$~ms, which is about one-tenth of the average impact duration observed in our experiments. The material-dependent COR is estimated to be $0.627$\footnote{\url{https://gite.lirmm.fr/yuquan/fidynamics/-/wikis/Estimating-the-material-dependent-coefficient-of-restitution.}}.
We applied the following reference contact velocities: $0.08$\unitVelTS, $0.10$\unitVelTS, $0.12$\unitVelTS, $0.15$\unitVelTS, $0.18$\unitVelTS for three distinct configurations, see Fig.~\ref{fig:impact_configurations}. We repeated 10 times each combination. Hence, the dataset includes  $3\times 5 \times 10 = 150$ experiments. 

We sample the force-torque sensor at $25$\unitkHz to capture the dynamics of  low-velocity impacts, see the 40 contact-force profiles in~\figRef{fig:measured_contact_force_profiles}. The Panda robot has a torque sensor on each joint. Once the impact is detected by thresholding the joint torques, the robot controller immediately pulls back the end-effector to avoid redundant post-impact actions. 
For the three impact configurations,
we noticed significant impact-induced joint torque errors associated with the 5th and 6th joint. Thus, we detect the impact by thresholding:
$$
\sum^{6}_{i = 5} \abs{\torque_i  - \desired{\torque}_i} \leq  \torque_0,
$$
where the QP controller updates the reference $\desired{\torque}_i$ at each control cycle. The threshold $\torque_0$ is $1.8$\unitTorque for configuration one and two; $1.5$\unitTorque for configuration three.

We have the ground-truth impact-timing from the force-torque sensor.
The measured forces do not suffer from motion-dependent drift as the sensor is rigidly attached to the wall.
We achieved $3$ to $5$\unitMs detection time, which is comparable to the  state-of-the-art collision detection time: $3$\unitMs~\cite{birjandi2020ral}.

\section{Experimental Results Dicussion}
\label{sec:validation}
Confronting the models to the obtained measurements, we found the followings:\\
{\bf c1}: only the nonlinear viscoelastic model regenerates measured contact forces and theory-consistent COR; see \secRef{sec:validate-contact-force-model}. \\
{\bf c2}: the proposed IIM computation \eqref{eq:iim_inertia_def} is the most accurate; see \secRef{sec:validate_iim}.

According to {\bf c1}, we claim the following remarks: \\
{\bf r1}: the pure material-dependent COR does not apply for high-stiffness controlled manipulators, see \figRef{fig:coe_estimation_parallel_connection}, 
\\
{\bf r2}: Since the \emph{virtual spring model} is not adequate, the following assumptions that are applied in~\cite{jia2017ijrr,jia2019ijrr} are not applicable for robot impacts: 
(1) the potential energy $\pEnergy$  reaches the peak at maximum contact force;
(2) the compression restores to zero when the restitution ends.

To verify if the impact (or restitution) ends, we check if the contact force~\eqref{eq:normal_impact_force} restores to zero.
We also observed that if the conditions  in~\secRef{sec:inelasticity} are met, we can apply a COR smaller than $0.15$  for our experiment setup. 


\subsection{Contact force model}
\label{sec:validate-contact-force-model}
The virtual-spring model does not fit measured contact-force profiles unless the COR is greater than~1, see~\secRef{sec:virtual_spring_model}). The Maxwell model assumes that the COR remains unchanged regardless of the contact velocities. However, this assumption contradicts the data, see~\secRef{sec:maxwell_model}. The candidate nonlinear viscoelastic model can match the measured contact-force profile with a theory-consistent COR, see~\secRef{sec:nonlinear_model}.

\subsubsection{The virtual spring model}
\label{sec:virtual_spring_model}
\begin{figure}[hbtp]
  \begin{tabular}{C{.22\textwidth}C{.22\textwidth}}
    \subfigure[b][The virtual spring model] {
    \resizebox{0.22\textwidth}{!}{%
    \begin{tikzpicture}
      \label{fig:virtual_spring}
      \begin{scope}
        \node[anchor=south west,inner sep=0] (image) at (0,0) {
          \includegraphics[width=\textwidth]{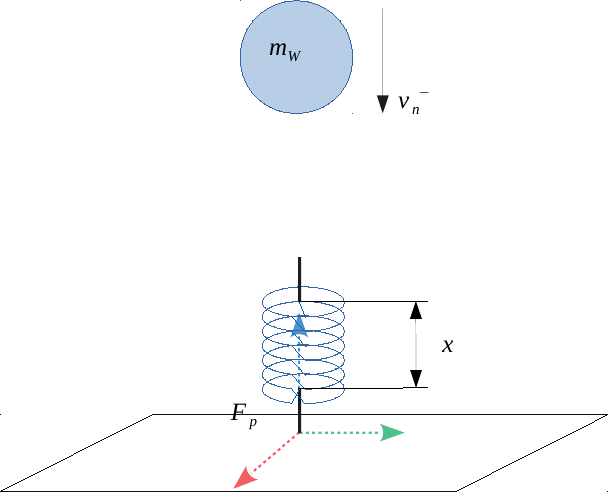}};
      \end{scope}
    \end{tikzpicture} 
    }
    }\subfigure[b][Requires COR greater than 1 ] {
    \resizebox{0.22\textwidth}{!}{%
    \begin{tikzpicture}
      \label{fig:spring_model_deficiency}
      \begin{scope}
        \node[anchor=south west,inner sep=0] (image) at (0,0) {
          \includegraphics[width=\textwidth]{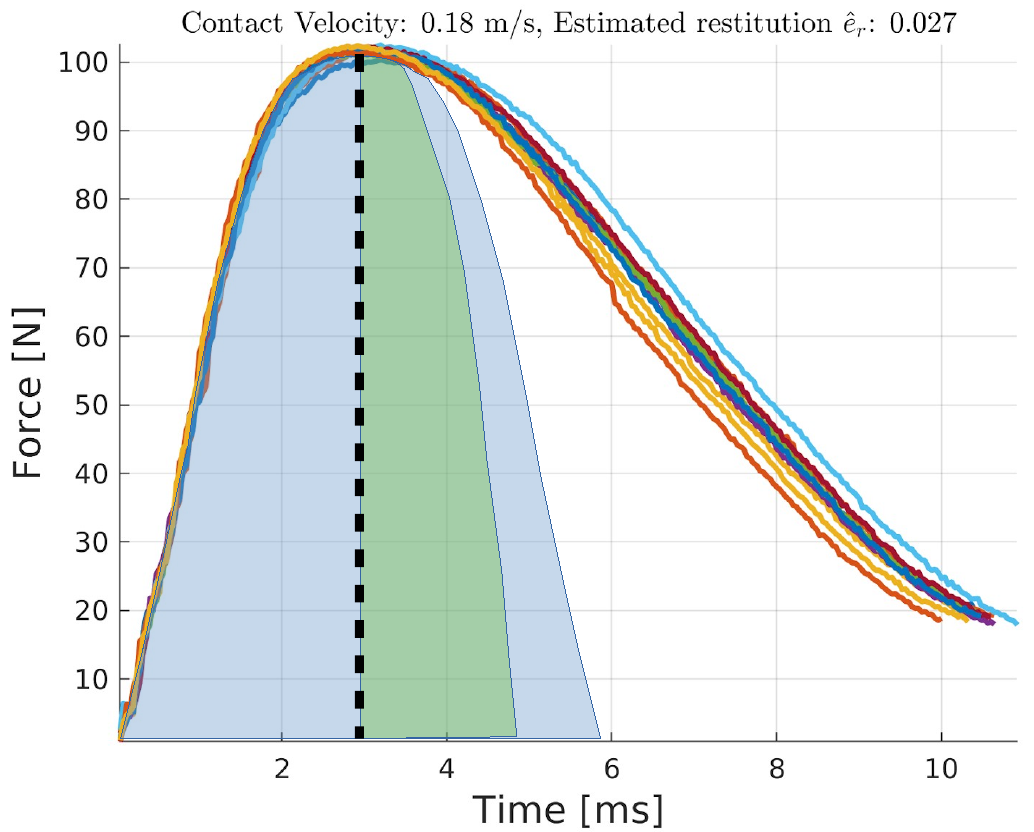}};
      \end{scope}
    \end{tikzpicture} 
    }
    }
  \end{tabular}
\caption{Fig.~\ref{fig:virtual_spring}: The virtual spring model. The light blue color in Fig.~\ref{fig:spring_model_deficiency} illustrates the area covered by the contact-force profile when $\eCoefR =1$. The green area is an example when $0<\eCoefR<1$. If $\eCoefR =0$, the contact-force profile ends at its maximum marked with the dashed vertical line.}
\label{fig:simulation_post-impact-states}
\end{figure}

According to the virtual spring in \figRef{fig:virtual_spring}, the contact force and the compression should be perfectly in phase; that is to say, the maximum contact force is reached when the compression ends.
Therefore, the contact-force profile during restitution should be within the blue area in \figRef{fig:spring_model_deficiency} according to $0<\eCoefR < 1$ \cite[Chapter~2.2]{stronge2000book}. Yet, the virtual-spring model does not fit the measured contact profiles unless the coefficient of restitution $\eCoefR > 1$! This observation is not limited to \figRef{fig:spring_model_deficiency}, and similar patterns are found in other contact-force profiles shown in~\figRef{fig:measured_contact_force_profiles}.

Thus, the virtual spring model is not suitable. The contact force shall be out-of-phase with the compression by some angle; i.e., the peak contact force occurs ahead of the maximum compression.
\subsubsection{The Maxwell model}
\label{sec:maxwell_model}
\begin{figure}[hbtp]
  \begin{tabular}{C{.22\textwidth}C{.22\textwidth}}
      \subfigure[b][Maxwell contact-force model] {
      \resizebox{0.22\textwidth}{!}{%
      \begin{tikzpicture}
        \label{fig:series_model_illustration}
        \begin{scope}
          \node[anchor=south west,inner sep=0] (image) at (0,0) {
            \includegraphics[width=\textwidth]{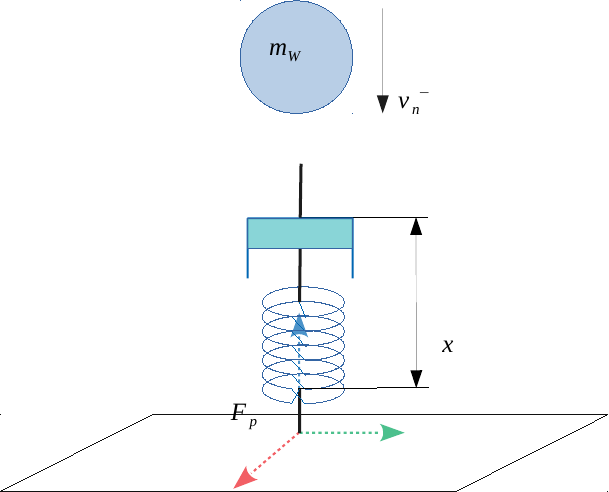}};
        \end{scope}
      \end{tikzpicture} 
      }
      }\subfigure[b][Contact velocity dependent COR] {
      \resizebox{0.22\textwidth}{!}{%
      \begin{tikzpicture}
        \label{fig:coe_estimation_series_connection}
        \begin{scope}
          \node[anchor=south west,inner sep=0] (image) at (0,0) {
            \includegraphics[width=\textwidth]{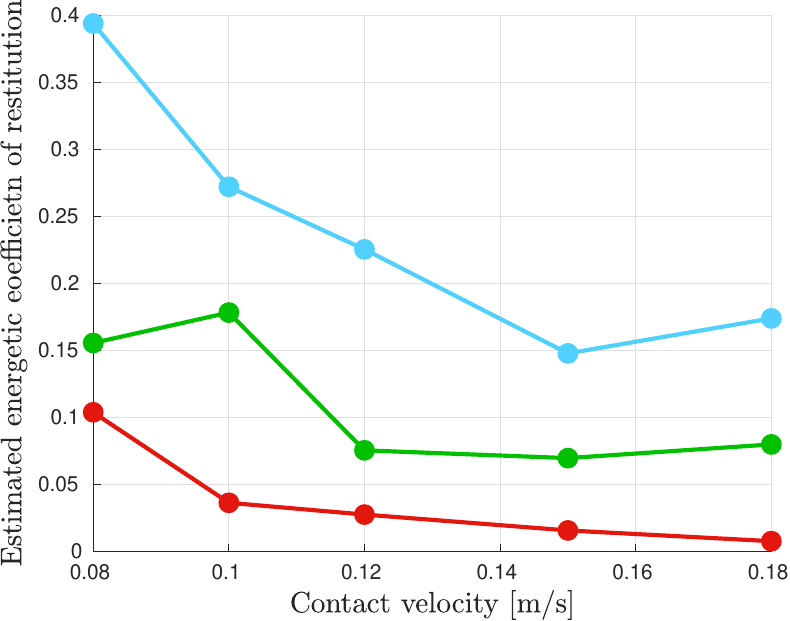}};
        \end{scope}
      \end{tikzpicture} 
      }
      }
  \end{tabular}
\caption{
  Fig.~\ref{fig:series_model_illustration}: Series connection of a spring and dashpot model. For the three impact configurations in Fig.~\ref{fig:impact_one},~\ref{fig:impact_two}, and~\ref{fig:impact_three},
  the  estimated COR in Fig.~\ref{fig:coe_estimation_series_connection} (red, blue, and green for one, two, and three) vary with respect to the increasing contact velocities.
  This is not in agreement with the assumption that the COR should be constant for such a model structure~\cite[Chapter~5.1.1]{stronge2000book}.
}
\label{fig:simulation_post-impact-states}
\end{figure}

We illustrate the Maxwell model (series connection of a spring with stiffness $\vSpringC$ and a dashpot with constant $\vDamperC$), see~\figRef{fig:series_model_illustration}. The complete  linear second-order system is detailed in~\cite[Chapter~5.1.1]{stronge2000book}. Measured contact-force profiles are exploited to identify $\vSpringC$,  $\vDamperC$, and COR according to~\cite[Eq~5.7]{stronge2000book}, using MATLAB estimation toolbox. The estimated COR should be invariant with respect to the increasing contact velocities. However, according to \figRef{fig:coe_estimation_series_connection}, this is not the case. In this case, the Maxwell model is also not suitable. 

\subsubsection{The nonlinear viscoelastic model}
\label{sec:nonlinear_model}
Similarly to~\secRef{sec:maxwell_model}, we identified $\vSpringC $ and $\vDamperC$ for the nonlinear viscoelastic model~\eqref{eq:normal_impact_force}; see~\cite[Chapter~5.1.2]{stronge2000book}. 

We overlay the model-generated contact-force profiles on the measurements in \figRef{fig:estimation}. Because of the viscoelasticity, the compression $x$ might not fully restore to zero by the end of the restitution phase.

\begin{figure}[hbtp]
  \begin{tabular}{C{.21\textwidth}C{.22\textwidth}}
      \subfigure[b][ The contact-force model] {
      \resizebox{0.21\textwidth}{!}{%
      \begin{tikzpicture}
        \label{fig:model_illustration}
        \begin{scope}
          \node[anchor=south west,inner sep=0] (image) at (0,0) {
            \includegraphics[width=\textwidth]{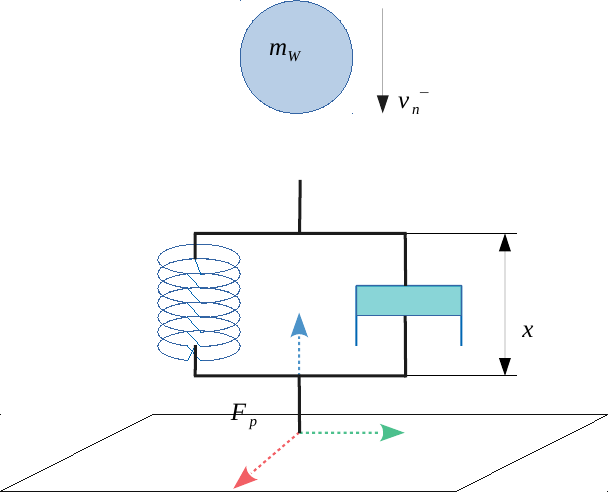}};
        \end{scope}
      \end{tikzpicture} 
      }
    } &
    \subfigure[b][ Theory-consistent COR] {
      \resizebox{0.22\textwidth}{!}{%
      \begin{tikzpicture}
        \label{fig:coe_estimation_parallel_connection}
        \begin{scope}
          \node[anchor=south west,inner sep=0] (image) at (0,0) {
            \includegraphics[width=\textwidth]{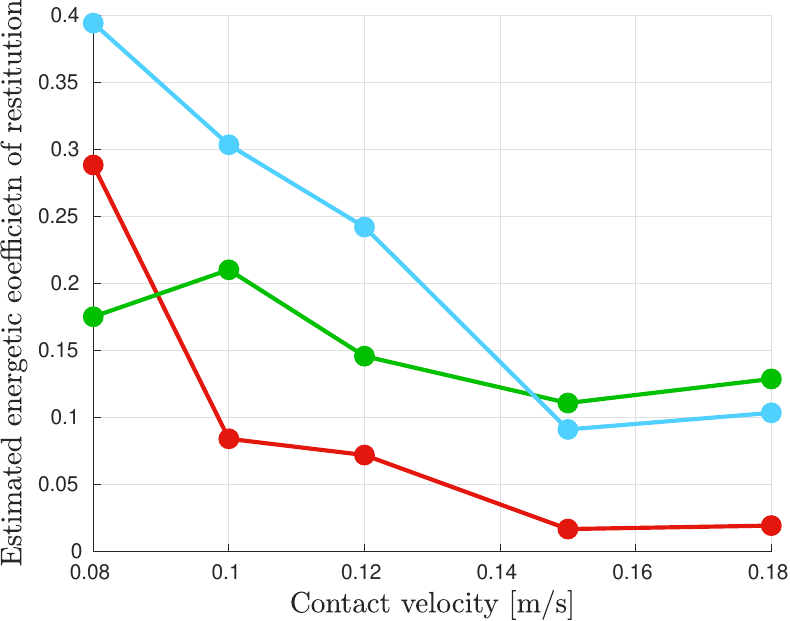}};
        \end{scope}
      \end{tikzpicture} 
      }
      }
  \end{tabular}
  \caption{
    The parallel connection of a spring and a dashpot model. For the three impact configurations in Fig.~\ref{fig:impact_one}, \ref{fig:impact_two}, and \ref{fig:impact_three}, the estimated COR in Fig.~\ref{fig:coe_estimation_parallel_connection}  (red, blue, and green for one, two, and three) roughly decreases while the contact velocity increases. This observation agrees in large with the analysis in Sec.~\ref{sec:cor_analysis} and \cite[Chapter~5.1.2]{stronge2000book}.
}
\label{fig:simulation_post-impact-states}
\end{figure}

\begin{figure*}[hbtp]
  \vspace{-3mm}
  \begin{tabular}{C{.48\textwidth}C{.48\textwidth}}
    \subfigure{
    \resizebox{0.24\textwidth}{!}{%
    \begin{tikzpicture}
      \begin{scope}
        \node[anchor=south west,inner sep=0] (image) at (0,0) {
          \includegraphics[width=\textwidth]{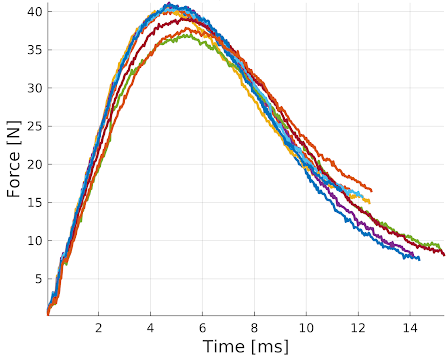}};
      \end{scope}
    \end{tikzpicture} 
    }
    }
         \subfigure{
         \resizebox{0.24\textwidth}{!}{%
         \begin{tikzpicture}
           \begin{scope}
             \node[anchor=south west,inner sep=0] (image) at (0,0) {
               \includegraphics[width=\textwidth]{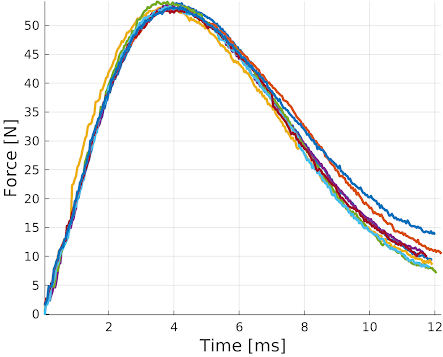}};
           \end{scope}
         \end{tikzpicture} 
         }
         }
    & 
          \subfigure{
    \resizebox{0.24\textwidth}{!}{%
    \begin{tikzpicture}
      \begin{scope}
        \node[anchor=south west,inner sep=0] (image) at (0,0) {
          \includegraphics[width=\textwidth]{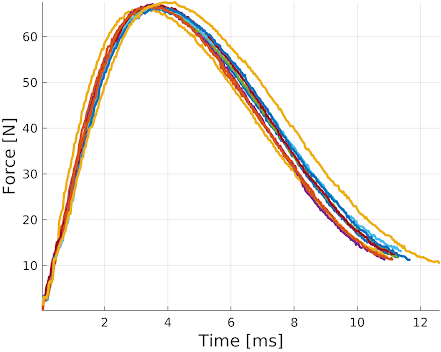}};
      \end{scope}
    \end{tikzpicture} 
    }
    }
         \subfigure{
         \resizebox{0.24\textwidth}{!}{%
         \begin{tikzpicture}
           \begin{scope}
             \node[anchor=south west,inner sep=0] (image) at (0,0) {
               \includegraphics[width=\textwidth]{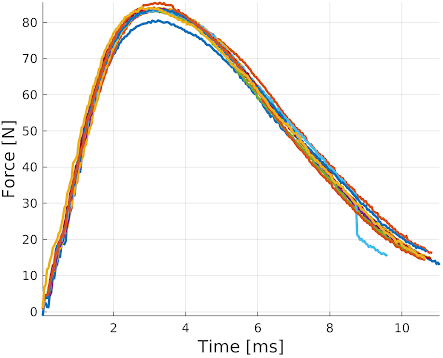}};
           \end{scope}
         \end{tikzpicture} 
         }
         }
\end{tabular}
\caption{
  From left to right, we plot four sets of  contact-force profiles. Each set includes data from 10 experiments. The corresponding contact velocities are:
  $0.0755$\unitVelTS, $0.0955$\unitVelTS, $0.1154$\unitVelTS, and $0.1455$\unitVelTS. The robot configuration at the impact time is shown in Fig.~\ref{fig:impact_one}. 
}
\label{fig:measured_contact_force_profiles}
\vspace{-3mm}
\end{figure*}

\begin{figure*}[htbp]
  \vspace{-5mm}
  \begin{tabular}{C{1\textwidth}}
    \subfigure{
    \resizebox{0.30\textwidth}{!}{%
    \begin{tikzpicture}
      \label{fig:es1}
      \begin{scope}
        \node[anchor=south west,inner sep=0] (image) at (0,0) {
          \includegraphics[width=\textwidth]{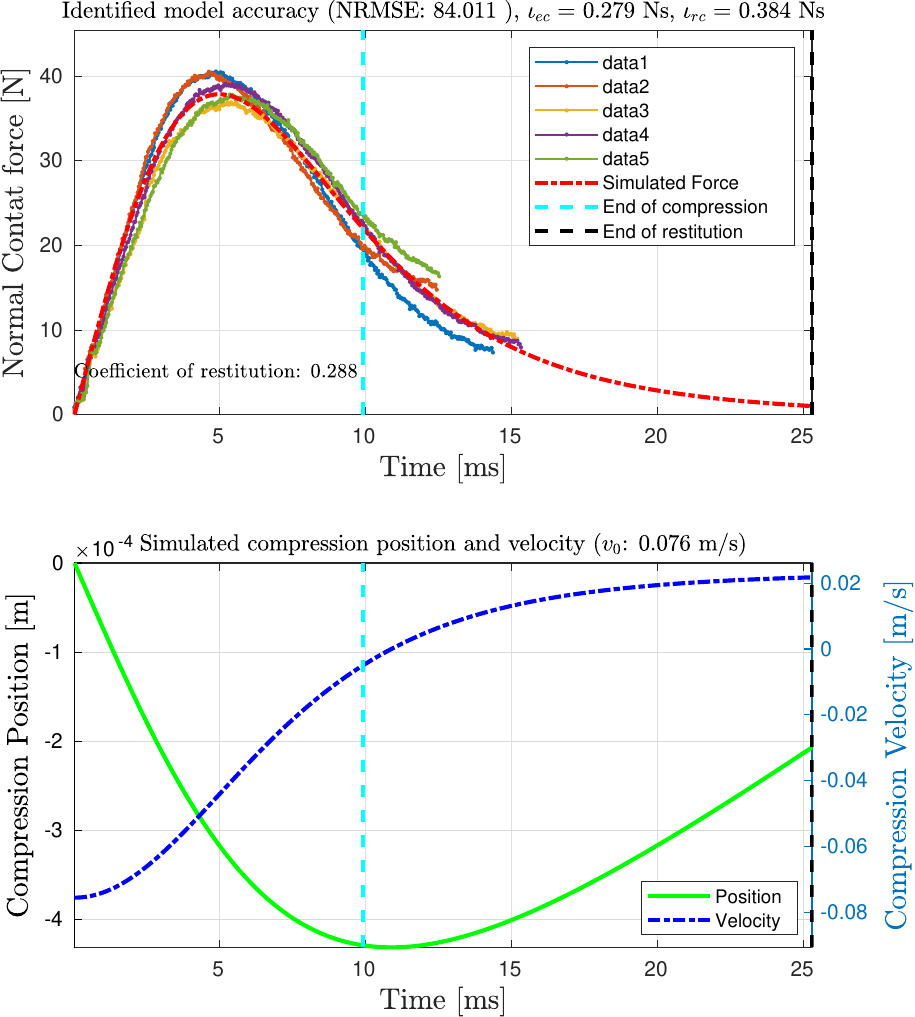}};
      \end{scope}
    \end{tikzpicture} 
    }
    }
    \subfigure{
    \resizebox{0.30\textwidth}{!}{%
    \begin{tikzpicture}
      \label{fig:es2}
      \begin{scope}
        \node[anchor=south west,inner sep=0] (image) at (0,0) {
          \includegraphics[width=\textwidth]{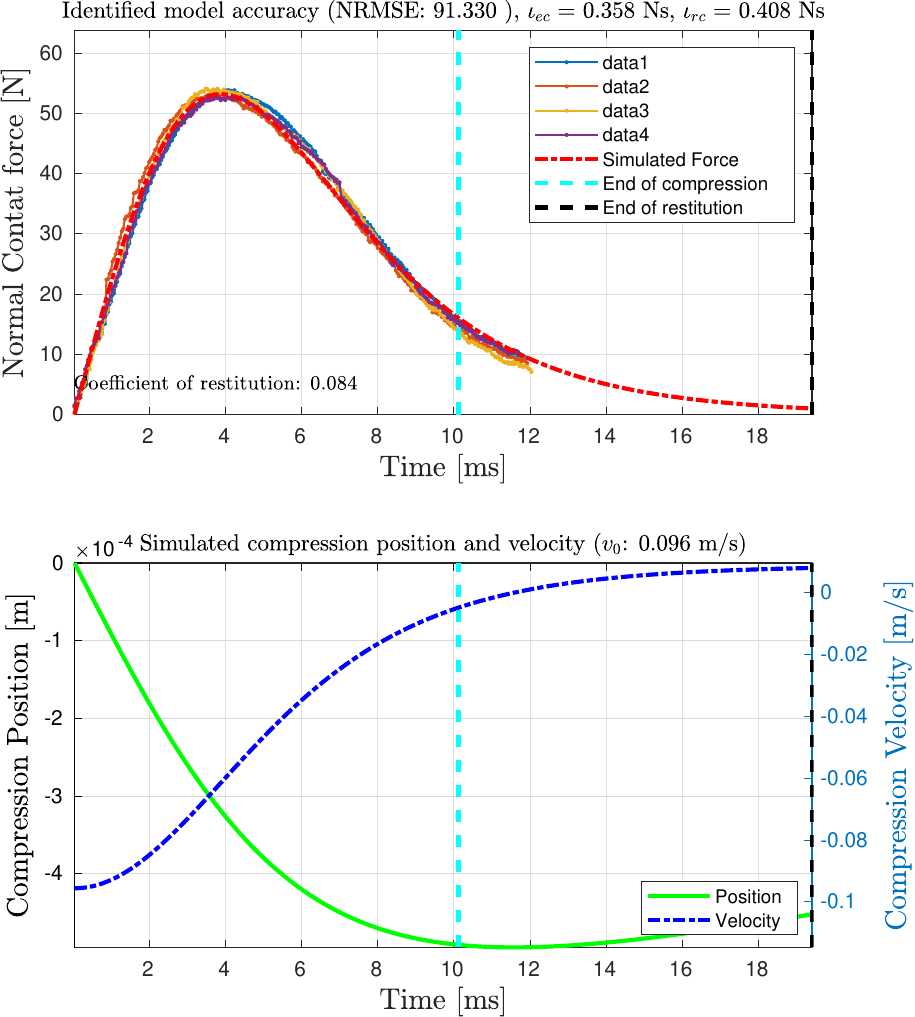}};
      \end{scope}
    \end{tikzpicture} 
    }
    }
    \subfigure{
    \resizebox{0.30\textwidth}{!}{%
    \begin{tikzpicture}
      \label{fig:es3}
      \begin{scope}
        \node[anchor=south west,inner sep=0] (image) at (0,0) {
          \includegraphics[width=\textwidth]{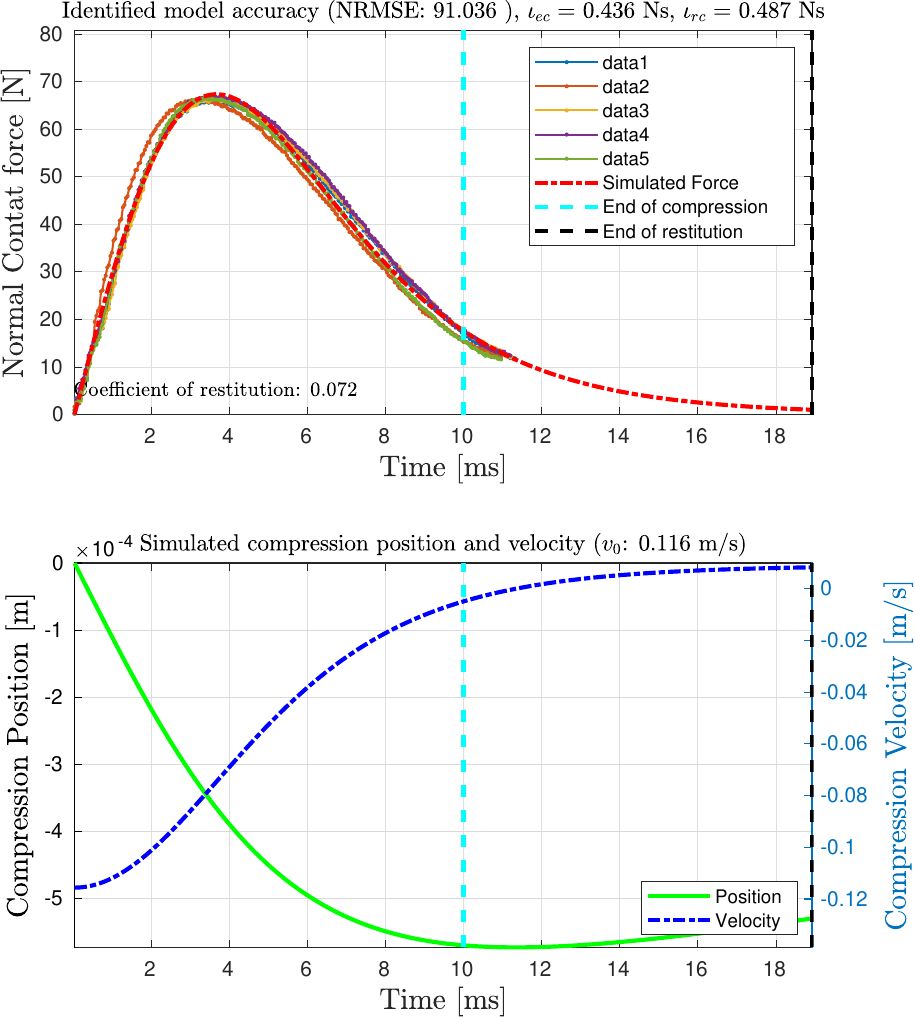}};
      \end{scope}
    \end{tikzpicture} 
    }
    }
  \end{tabular}
  \caption{
    We overlay the contact-force generated from the nonlinear viscoelastic model \eqref{eq:normal_impact_force} on the measurements. From left to right, the contact velocities are $0.0755$, $0.0955$, and $0.1154$\unitVelTS. We mark the moments when: (1) the compression ends, i.e., the compression rate is zero $\dot{x} = 0$; and (2) the restitution ends, i.e., the contact-force  is zero $\nForce = 0$. Through the impact process, the $\dot{x}$ increases monotonically from the negative initial value. The compression does not systematically resume to zero. 
  }
  \label{fig:estimation}
  \vspace{-5mm}
\end{figure*}

According to~\eqref{eq:coe_property}, the COR should decrease while the contact velocity increases, which is mostly true in \figRef{fig:coe_estimation_parallel_connection}. The exception happened at the second impact configuration when the reference contact velocity is $0.08$\unitVelTS.

\begin{figure*}[htbp]
  \vspace{-3mm}
  \begin{tabular}{C{1\textwidth}}
    \subfigure{
    \resizebox{0.30\textwidth}{!}{%
    \begin{tikzpicture}
      \label{fig:energy1}
      \begin{scope}
        \node[anchor=south west,inner sep=0] (image) at (0,0) {
          \includegraphics[width=\textwidth]{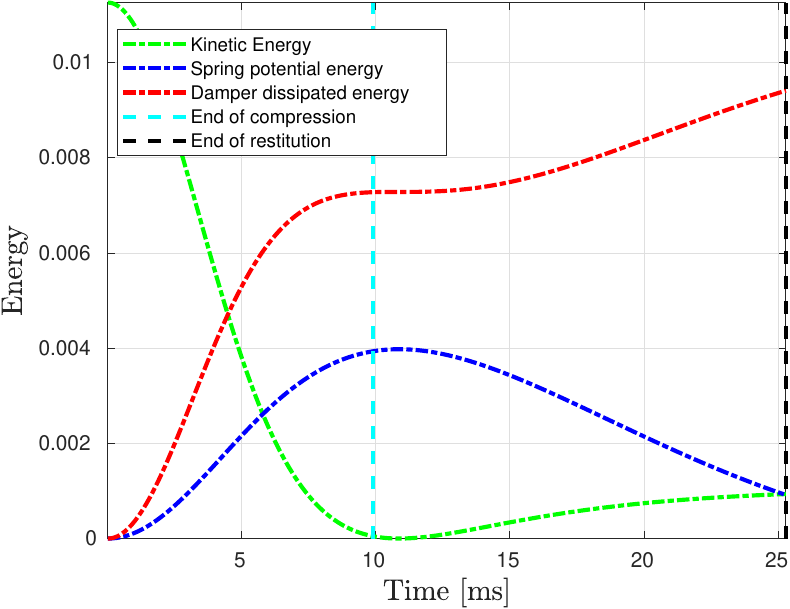}};
      \end{scope}
    \end{tikzpicture} 
    }
    }
    \subfigure{
    \resizebox{0.30\textwidth}{!}{%
    \begin{tikzpicture}
      \label{fig:energy2}
      \begin{scope}
        \node[anchor=south west,inner sep=0] (image) at (0,0) {
          \includegraphics[width=\textwidth]{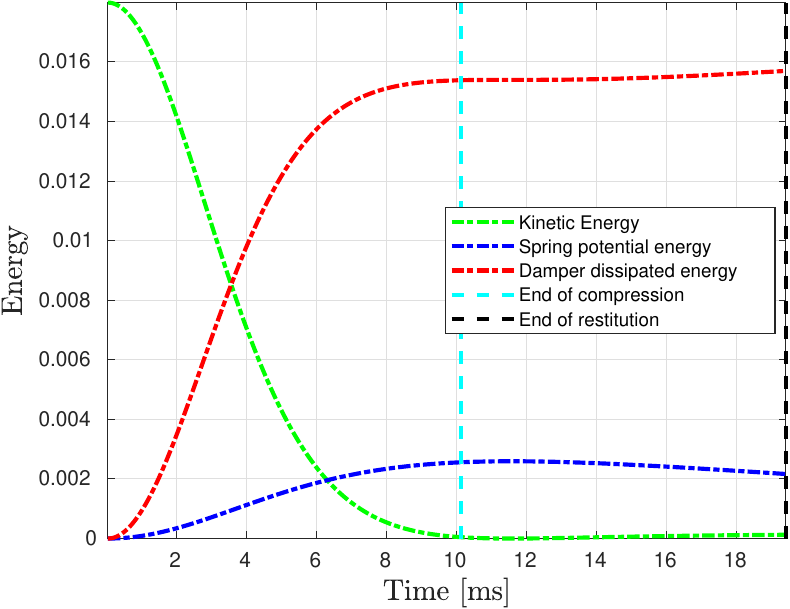}};
      \end{scope}
    \end{tikzpicture} 
    }
    }
    \subfigure{
    \resizebox{0.30\textwidth}{!}{%
    \begin{tikzpicture}
      \label{fig:energy3}
      \begin{scope}
        \node[anchor=south west,inner sep=0] (image) at (0,0) {
          \includegraphics[width=\textwidth]{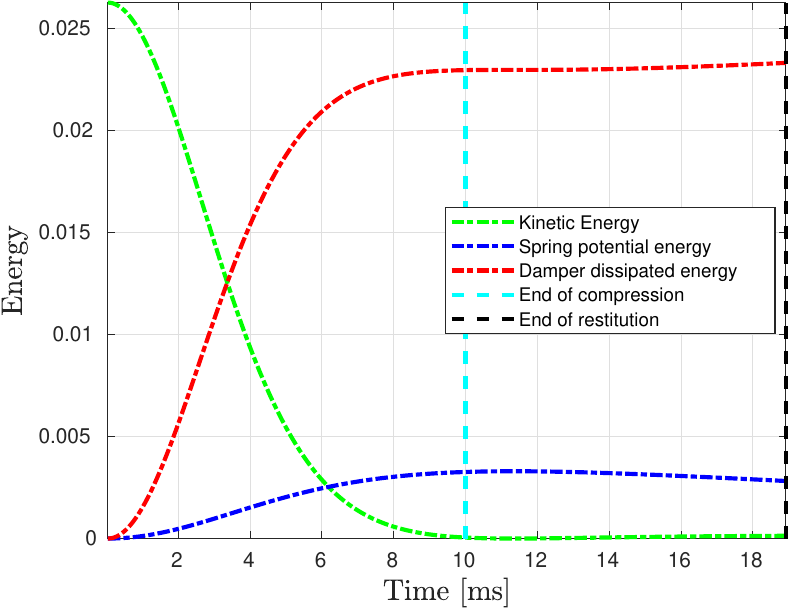}};
      \end{scope}
    \end{tikzpicture} 
    }
    }
  \end{tabular}
  \caption{
    The sum of the
    potential energy \eqref{eq:potential_energy},
    the dissipated energy \eqref{eq:dissipated_energy} , and
    the kinetic energy
    is equal to the initial kinetic energy: $\frac{1}{2}\mass {\preImpact{\nVel}}^2$, see \eqref{eq:energy_balance}.
  }
  \label{fig:energy}
  \vspace{-3mm}
\end{figure*}

The spring model in~\figRef{fig:virtual_spring} cannot capture the energy dissipation. It entirely relies on the COR to model the energy loss. For instance, it relies on a sudden decrease of the potential energy by $\eCoefR^2 \frac{1}{2}\mass {\preImpact{\nVel}}^2$ when the compression phase ends~\cite[Eq.~11, Fig.~3.(h)]{jia2019ijrr}. Due to the dashpot, we can explicitly and continuously describe the potential, kinetic, and dissipated energy along with their derivatives; see the equations in \secRef{sec:energy} and the plots in~\figRef{fig:energy}. 

\subsection{Candidate inverse inertia matrices}
\label{sec:validate_iim}

\begin{figure*}[htp!]
  \vspace{-3mm}
  \begin{tabular}{C{1\textwidth}}
    \subfigure[b][Impulses corresponding to  Fig.~\ref{fig:impact_one}] {
    \resizebox{0.30\textwidth}{!}{%
    \begin{tikzpicture}
      \label{fig:em1}
      \begin{scope}
        \node[anchor=south west,inner sep=0] (image) at (0,0) {
          \includegraphics[width=\textwidth]{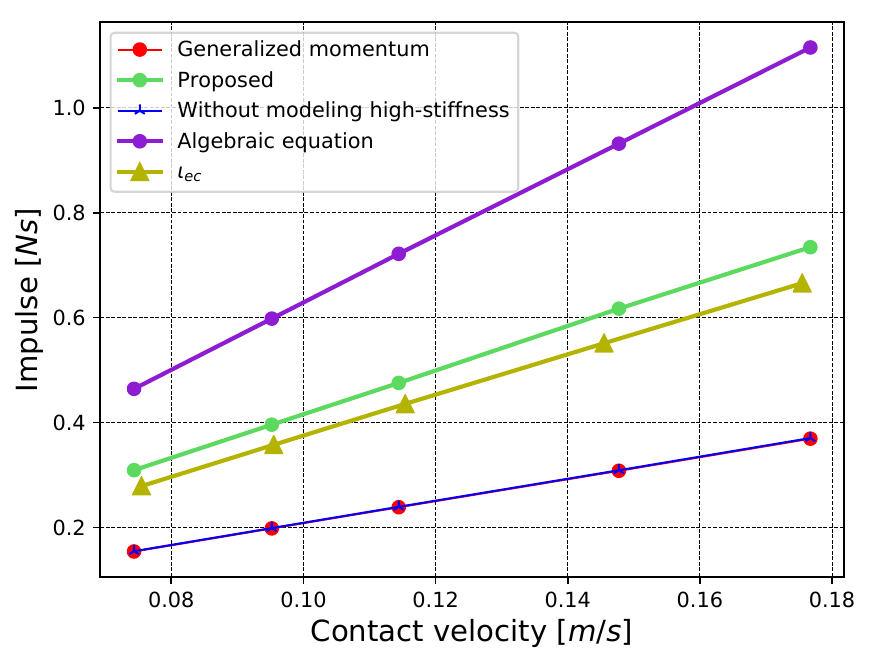}};
      \end{scope}
    \end{tikzpicture} 
    }
    }
    \subfigure[b][Impulses corresponding to  Fig.~\ref{fig:impact_two}] {
    \resizebox{0.30\textwidth}{!}{%
    \begin{tikzpicture}
      \label{fig:em2}
      \begin{scope}
        \node[anchor=south west,inner sep=0] (image) at (0,0) {
          \includegraphics[width=\textwidth]{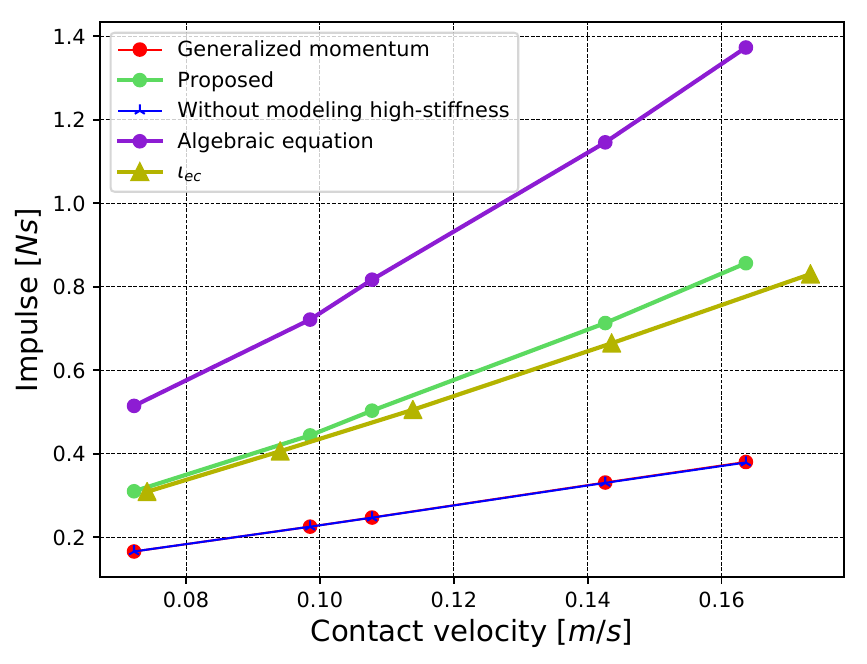}};
      \end{scope}
    \end{tikzpicture} 
    }
    }    
    \subfigure[b][Impulses corresponding to  Fig.~\ref{fig:impact_three}] {
    \resizebox{0.30\textwidth}{!}{%
    \begin{tikzpicture}
      \label{fig:em3}
      \begin{scope}
        \node[anchor=south west,inner sep=0] (image) at (0,0) {
          \includegraphics[width=\textwidth]{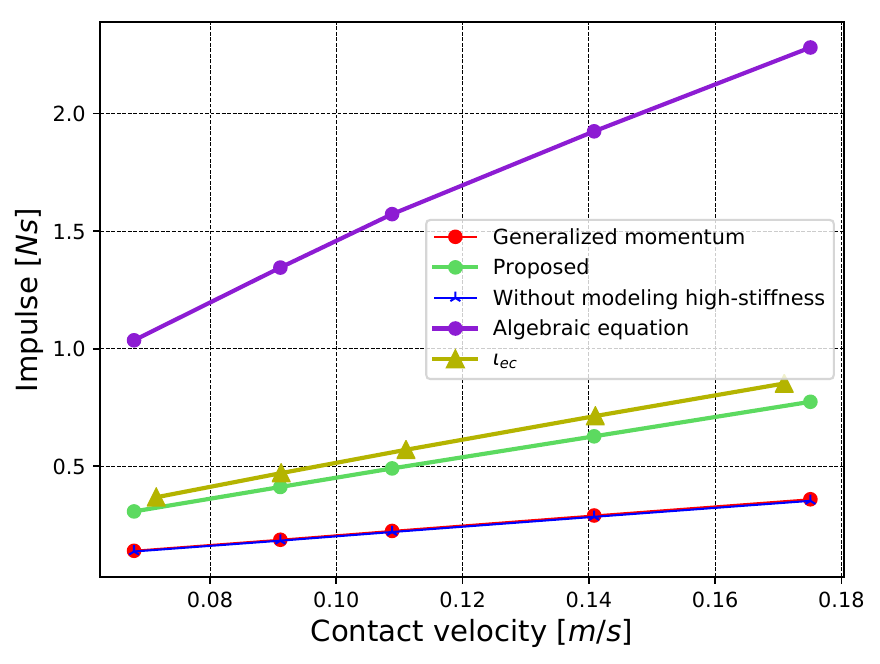}};
      \end{scope}
    \end{tikzpicture} 
    }
    }
  \end{tabular}
  \caption{
    We compare various predicted end-of-compression impulse $\nImpulseCE$  against the measurements.
    We compute  each prediction with the mean joint configuration of  10 experiments.
    The CRB approach \eqref{eq:iim_inertia_def} is close to the mean of the measured impulse (the yellow triangles).  The algebraic equation \eqref{eq:sota_gm_iim}  overestimates, while option \eqref{eq:crb_fb_iim} and the generalized momentum approach \eqref{eq:option_gm_iim} underestimates.
  }
  \label{fig:em}
  \vspace{-3mm}
\end{figure*}

We compute the impulse\footnote{The impulse $\nImpulseCE$ does not rely on COR as it corresponds to the moment when the compression ends.}  $\nImpulseCE$ using different options: 
\begin{enumerate}
\item \label{opt:sota} the algebraic equation~\eqref{eq:sota_gm_iim}, see \cite{grizzle2014automatica,siciliano2016springer,aouaj2021icra,zheng1985mathematical};
\item \label{opt:inverse_gm} the generalized momentum approach \cite{khulief2013modeling,lankarani2000poisson}: substituting $\iim_{\text{gm}}$ \eqref{eq:option_gm_iim} into \eqref{eq:zImpulseCE};
\item \label{opt:crb} the CRB approach: substituting $\iim$ \eqref{eq:iim_inertia_def} into \eqref{eq:zImpulseCE};
\item \label{opt:crb_fb} without considering joints' high-stiffness: substituting $\tilde{\iim}$ \eqref{eq:crb_fb_iim} into \eqref{eq:zImpulseCE}.
\end{enumerate}
Option~\ref{opt:sota} overestimates in all the situations as shown in~\figRef{fig:em}.
Hence, computation according to kinetic energy conservation in the joint space~\eqref{eq:kinetic_energy_conservation} is not a good hypothesis. 
Options~\ref{opt:inverse_gm} and~\ref{opt:crb_fb} lead to similar results: both showing underestimated performance. Thus, we cannot assume the joints are completely flexible in the joint motion subspace, as in the under-actuated pendulum by Lankarani~\cite{lankarani2000poisson} or the under-actuated linkage by Stronge~\cite[Example~8.1]{stronge2000book}.

Option~\ref{opt:crb} leads to the most accurate prediction. Therefore, the CRB assumption applies to a high-stiffness-controlled manipulator.

\subsection{Small coefficient of restitution}
\label{sec:inelasticity}

\begin{table}[htbp!]
  \caption{
    We compare the contact velocity \eqref{eq:contactVel_exact} to its approximation \eqref{eq:contactVel_transform}. }
\begin{tabular}{l*{4}{c}r}
  Reference: & 0.08~m/s\hspace{-3mm} & 0.10~m/s\hspace{-3mm} & 0.12~m/s\hspace{-3mm} & 0.15~m/s\hspace{-3mm} & 0.18~m/s\hspace{-3mm}    \\
  \hline
  \multicolumn{6}{c}{Exact: $\innerP{\basisVec{n}}{\bodyTV{\inertialFrame}{\contactPoint}}$. }\\
  \hline
  Fig.~\ref{fig:impact_one}    &0.0756 & 0.0956 & 0.1156& 0.1456 & 0.1755\\
  Fig.~\ref{fig:impact_two} &0.0745 & 0.0943 & 0.1141 & 0.1438 & 0.1734\\
  Fig.~\ref{fig:impact_three} &0.0715& 0.0913& 0.1111& 0.1411& 0.1709\\    
  \hline
  \multicolumn{6}{c}{Approximation: $\innerP{\basisVec{n}}{(\bodyTV{\inertialFrame}{\contactPoint} - \bodyTV{\com}{\contactPoint})}$. }\\
  \hline
  Fig.~\ref{fig:impact_one}    &0.0699 & 0.0870 & 0.1056 & 0.1309 & 0.1623 \\
  Fig.~\ref{fig:impact_two} & 0.0567 & 0.0734 & 0.0975 &0.1239 & 0.1486\\
  Fig.~\ref{fig:impact_three}    &0.0446 &0.0586 & 0.0712 & 0.0894 & 0.1036\\
  \hline
  \multicolumn{6}{c}{Approximate-to-exact ratio: $\coefA$.
    }\\
  \hline
  Fig.~\ref{fig:impact_one}   &0.9248    &0.9100    &0.9141    &0.8996    &0.9243\\
  Fig.~\ref{fig:impact_two}     &0.7609    &0.7785    &0.8544    &0.8618    &0.8567\\
  Fig.~\ref{fig:impact_three}   &0.6244    &0.6423    &0.6406    &0.6336    &0.6062\\
\end{tabular}
\label{tab:relative_vels}
\vspace{-3mm}
\end{table}

Given the linear relative velocity $\bodyTV{\com}{\contactPoint}$ 
and the impact normal 
$\basisVec{n} \in \RRv{3}$, we define the approximate-to-exact  ratio as:
\quickEq{eq:approximate-to-exact}{
  \coefA = \frac{
    \innerP{\basisVec{n}}{(\bodyTV{\inertialFrame}{\contactPoint} - \bodyTV{\com}{\contactPoint})}
  }
  {
    \innerP{\basisVec{n}}{\bodyTV{\inertialFrame}{\contactPoint}}
  }.
}
The smaller the projection $\innerP{\basisVec{n}}{\bodyTV{\com}{\contactPoint}}$, the higher  $\coefA$.

According to the numerical values in Table~\ref{tab:relative_vels} and the corresponding estimated COR in~\figRef{fig:coe_estimation_parallel_connection}, we observe that it is possible to assume $\coefR < 0.15$ (i.e., approximately inelastic impact) in our experiments if the following are met: \\
(1) when  the relative velocity is close to zero, e.g.,  when the approximate-to-exact ratio $\coefA > 0.85$;\\
(2) the contact velocity is greater than $0.1$\unitVelTS.

\section{Conclusion}
Our objective is to devise an impact-aware controller for high-stiffness controlled robotic manipulators based on models, whether physics-driven or data-driven.
Our dataset of 150 impact experiments revealed the short-comings of the most-used models. Thus, we revisit two main ingredients:
\begin{itemize}
\item the inverse inertia matrix (IIM) computation that determines the effective mass (velocity-to-impulse mapping) at the impact and post-impact contact modes, and
\item the contact-force model that determines the impact event timing, namely that of the restitution phase. 
\end{itemize}

Our findings suggest that
\begin{enumerate}
\item one shall compute the inverse inertia matrix (IIM) as the inverse of the composite-rigid-body inertia transformed at the contact point, see~\secRef{sec:crb};
\item the widely-used virtual spring model~\cite{jia2017ijrr,khulief2013modeling} does not reproduce the impact behavior of high-stiffness controlled robots; 
\item the viscoelastic contact-force model (parallel connection of a virtual spring and a dashpot) in~\secRef{sec:contact_force_model} matches the measurements while fulfilling its assumptions. 
\end{enumerate}

In future work, we aim at accurately predicting the post-impact states when the contact surface is frictional and the tangential contact velocity is significant.
 

\begin{appendices}
  \section*{Appendix}
\subsection{Normal impulse computation}
\label{sec:nImpulse_calc}
The impact process consists of two sequential phases: compression and restitution \cite{jia2017ijrr}.
The compression ends when the normal contact velocity increases from $\preImpact{\nContactVel}<0$ to zero. To check the evolution of $\nContactVel$, we compute its derivative: 
$$
\dni{\nContactVel} = \dni{(\preImpact{\nContactVel} + \transpose{\basisVec{n}} \iim \impulse)} = \transpose{\basisVec{n}} \iim \dni{ \impulse} = \transpose{\basisVec{n}} \iim \basisVec{n} + \transpose{\basisVec{n}} \iim \dni{ \tp{\impulse}},
$$
where the vector $\basisVec{n} \in \RRv{3}$ denotes the  impact normal, the subscript $\tp{~}$ denotes the quantity is projected to the tangential plane.
Due to the assumption of a small friction coefficient, the tangential impulse is negligible. 
Thus, we have $\dni{\tp{\impulse}} = 0$. The contact velocity  $\nContactVel$ monotonically increases according to:
\quickEq{eq:v_z_d}{
  \dni{\nContactVel} = \dni{(\preImpact{\nContactVel} + \transpose{\basisVec{n}} \iim \impulse)} = \transpose{\basisVec{n}} \iim \dni{ \impulse} = \transpose{\basisVec{n}} \iim \basisVec{n} > 0,
}
which holds due to the positive definiteness of $\iim$ \cite{jia2019ijrr}. Integrating \eqref{eq:v_z_d} by separating $d\nContactVel$ and $d\nImpulse$ on two sides, the impulse writes:
\quickEq{eq:impulse_calc}{
\nImpulse = \emiim{\iim}(\nContactVel - \preImpact{\nVel}),
}
where we defined the positive scalar $\emiim{\iim} = \emiimDef{\iim}{\basisVec{n}}$. At the moment when $\nContactVel = 0$, we conclude the end of compression impulse:
\quickEq{eq:zImpulseCE}{
  \nImpulseCE = - \frac{\preImpact{\nContactVel}}{\transpose{\basisVec{n}} \iim \basisVec{n}}
  = - \emiim{\iim}\preImpact{\nContactVel}.
}

\subsection{Proof of remark \ref{remark:inertia}}
\label{sec:inertia_proof}
We prove \eqref{eq:contribution_im} by the  $\crbGInertia$  derivation in \appRef{app:ci}, and the analysis of a particular inertia matrix $\inertiaMatrix_i$ in \appRef{sec:i_contribution}.
\subsubsection{Centroidal inertia derivation}
\label{app:ci}
$\crbGInertia \in \RRm{6}{6}$ is identical to \cite[Eq.~22]{orin2013auro} from \eqref{eq:cm_jump}  as: 
\quickEq{eq:crb_i_derivation}{
\begin{aligned}
\jump \momentum & = \agg{i}{\nAJ}{
  \geometricFT{i}{\com}
  \jump \momentum_i}  =
\agg{i}{\nAJ}{
  \geometricFT{i}{\com}
  \inertiaMatrix_i
  \jump \bodyVel{\inertialFrame}{i}
}\\
& = \agg{i}{\nAJ}{
  \geometricFT{i}{\com}
  \inertiaMatrix_i
  \twistTransform{\com}{i}\twistTransform{i}{\com}\jump \bodyVel{\inertialFrame}{i}
} \\
&  =
\underbrace{
  [\geometricFT{1}{\com}, \ldots,  \geometricFT{n}{\com}]
\matrixThree{\inertiaMatrix_1}{\ldots}{0}
{\vdots}{\ddots}{\vdots}
{0}{\ldots}{\inertiaMatrix_n}
\vectorThree{\twistTransform{\com}{1}}{\vdots}{\twistTransform{\com}{n}}
}_{\crbGInertia}
\\
& [\twistTransform{1}{\com}, \ldots, \twistTransform{n}{\com}]
\vectorThree{\bodyVel{\inertialFrame}{1}}{\vdots}{\bodyVel{\inertialFrame}{n}}
\\
& =
\crbGInertia \agg{i}{\nAJ}{
  \twistTransform{i}{\com}\jump \bodyVel{\inertialFrame}{i}.
}
\end{aligned}
}
\subsubsection{Contribution of a particular inertia matrix}
\label{sec:i_contribution}
According to the derivation \eqref{eq:crb_i_derivation}, the \emph{contribution} of $\inertiaMatrix_i$ to $\crbGInertia$ is:
$$
\crbGInertia_i = \geometricFT{i}{\com} \inertiaMatrix_i \twistTransform{\com}{i}.
$$
Thus, employing the inertia transform \eqref{eq:m_transform} and the definition of $\inverse{\eqInertiaMatrix}_{\contactPoint}$ \eqref{eq:def_I_eq}, we obtain the \emph{contribution} of $\inertiaMatrix_i$ as: 
$$
\begin{aligned}
  &
\inverse{
  (\gInertiaTransform{\contactPoint}{\com}{
    \crbGInertia_i
})
} = 
  \inverse{
  (\gInertiaTransform{\contactPoint}{\com}{
  \geometricFT{i}{\com} \inertiaMatrix_i \twistTransform{\com}{i}
})
} \\
& =  \inverse{
  (
  \geometricFT{i}{\contactPoint} \inertiaMatrix_i \twistTransformTwo{\com}{i} \twistTransformTwo{\contactPoint}{\com}
  )}
= \inverse{
  (
  \geometricFT{i}{\contactPoint} \inertiaMatrix_i \twistTransformTwo{\contactPoint}{i}
  )} \\
& =
\twistTransformTwo{i}{\contactPoint}\inverse{\inertiaMatrix}_i \geometricFTTwo{\contactPoint}{i}.
\end{aligned}
$$


\end{appendices}
\section*{Acknowledgment}
We thank J.~Roux, P.~Gergondet, S.~Samadi, M.~Djeha and O. Tempier for helping in the experiments. We also thank our colleagues from the I.AM. consortium for their feedback.



\bibliographystyle{IEEEtran}
\bibliography{ref}

\end{document}